\pdfoutput=1
\documentclass{article}
\usepackage{multirow}
\usepackage{microtype}
\usepackage{graphicx}
\usepackage{booktabs} 

\usepackage{hyperref}
\usepackage{arydshln}

\usepackage[accepted]{icml2024}

\usepackage{amsmath}
\usepackage{amssymb}
\usepackage{mathtools}
\usepackage{amsthm}
\newcommand{\ourmeo}{\textbf{\texttt{CDTD}}}
\newcommand{\ourmeos}{\textbf{\texttt{CDTD}} }
\newcommand{\melater}{\ourmeos}

\usepackage{amsmath,amsfonts,bm}

\def\eqref#1{equation~\ref{#1}}

\def\1{\bm{1}}

\def\rvs{{\mathbf{s}}}

\def\rvu{{\mathbf{u}}}

\def\rvx{{\mathbf{x}}}

\def\rvz{{\mathbf{z}}}

\def\vtheta{{\bm{\theta}}}

\DeclareMathAlphabet{\mathsfit}{\encodingdefault}{\sfdefault}{m}{sl}
\SetMathAlphabet{\mathsfit}{bold}{\encodingdefault}{\sfdefault}{bx}{n}

\usepackage[capitalize,noabbrev]{cleveref}
\usepackage{wrapfig,lipsum,booktabs}
\usepackage{subfig}

\theoremstyle{plain}

\theoremstyle{definition}

\theoremstyle{remark}

\usepackage[textsize=tiny]{todonotes}

\newcommand{\bfsection}[1]{\noindent\textbf{#1.}}
\def\HS{\hspace{\fontdimen2\font}}
\definecolor{bananayellow}{rgb}{1.0, 0.88, 0.21}
\icmltitlerunning{Learning Causal Domain-Invariant Temporal Dynamics for Few-Shot Action Recognition}

\begin{document}

\twocolumn[


\icmltitle{Learning Causal Domain-Invariant Temporal Dynamics for \\ Few-Shot Action Recognition}

\icmlsetsymbol{equal}{*}

\begin{icmlauthorlist}
\icmlauthor{Yuke Li}{equal,yyy}
\icmlauthor{Guangyi Chen}{equal,comp}
\icmlauthor{Ben Abramowitz}{sch}
\icmlauthor{Stefano Anzellotti}{yyy}
\icmlauthor{Donglai Wei}{yyy}
\end{icmlauthorlist}

\icmlaffiliation{yyy}{Boston College, Boston MA, USA}
\icmlaffiliation{comp}{Carnegie Mellon University, Pittsburgh PA, USA; Mohamed bin Zayed University of Artificial Intelligence, Abu Dhabi, UAE}
\icmlaffiliation{sch}{Tulane University, New Orleans LA, USA}

\icmlcorrespondingauthor{Donglai Wei}{weidf@bc.edu}
\icmlcorrespondingauthor{Yuke Li}{sunfreshing@whu.edu.cn}

\icmlkeywords{Machine Learning, ICML}

\vskip 0.3in
]

\printAffiliationsAndNotice{\icmlEqualContribution} 

\begin{abstract}
    Few-shot action recognition aims at quickly adapting a pre-trained model to the novel data with a distribution shift using only a limited number of samples.
    Key challenges include how to identify and leverage the transferable knowledge learned by the pre-trained model.
    We therefore propose \ourmeo, or \underline{\textbf{C}}ausal \underline{\textbf{D}}omain-Invariant \underline{\textbf{Te}}mporal \underline{\textbf{D}}ynamics for knowledge transfer. 
    To identify the temporally invariant and variant representations, we employ the causal representation learning methods for unsupervised pertaining, and then tune the classifier with supervisions in next stage.      
    Specifically, we assume the domain information can be well estimated and the pre-trained image decoder and transition models can be well transferred. 
    During adaptation, we fix the transferable temporal dynamics and update the image encoder and domain estimator.
    The efficacy of our approach is revealed by the superior accuracy of \ourmeos over leading alternatives across standard few-shot action recognition datasets. 
\end{abstract}

\section{Introduction}

\newcommand{\ben}[1]{\textcolor{blue}{Ben says: #1}}


\begin{figure}[h]
\begin{center}
\includegraphics[width=\linewidth]{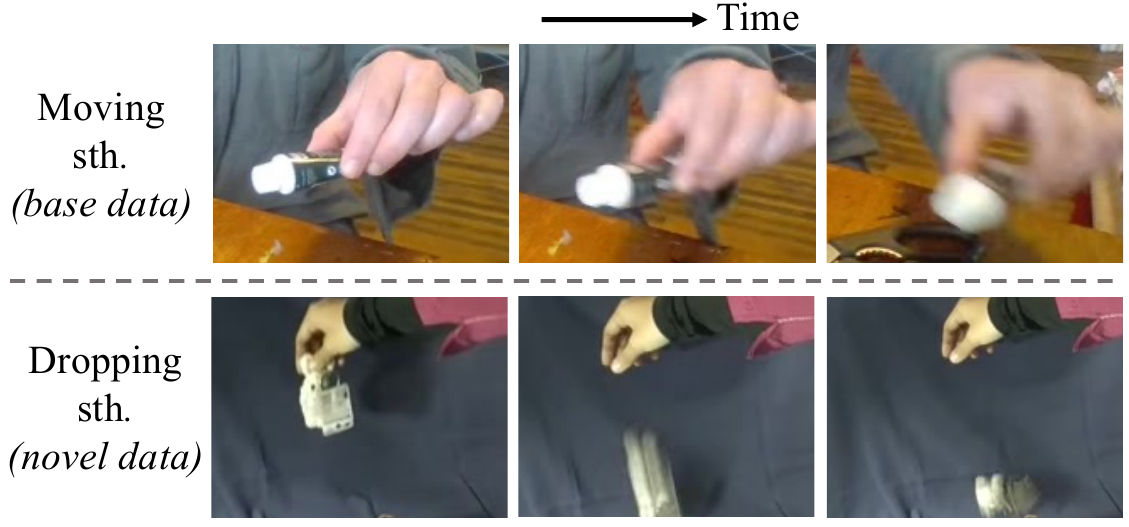}
\end{center}
\caption{ 
In the few-shot learning setting, what aspects can be effectively transferred from the base data to the novel data? Despite the different \textit{temporal dynamics} in these two videos, the underlying physical laws are \textit{domain-invariant}.}
\vspace{-0.4cm}
\label{fig:toy}
\end{figure}

Action recognition continues to be a potent and productive area of research. For instance, \cite{ac_cvpr23_1,ac_cvpr23_2,ac_cvpr23_3,ac_iccv23} show great results in learning action representations with a large amount of labels.
However, these supervised learning approaches may fall short when faced with the challenge of few-shot learning, where only a limited number of samples are available for novel action classes that exhibit significant distributional differences from the pre-training data \cite{few_cvpr20,few_cvpr21,few_cvpr22_1,fewshot_iclr20, few_icml23}. 


The efficient and effective few-shot action recognition requires solving two main problems. One entails identifying the transferable, temporal invariant knowledge that can be readily applied to new data. The other involves optimizing the non-transferable, temporal variant knowledge to facilitate swift updates, ensuring effective adaptation to new contexts. The existing literature relies on a large amount of labels and treats these two problems in various ways, based on different assumptions about transferability. For instance, one set of methods assumes all parameters should be fine-tuned on the new data, such as ActionCLIP \cite{ActionCLIP_arxiv21}, ORViT \cite{sthelse_cvpr22}, and SViT \cite{sthelse_neurips22}.
Other methods fix all parameters in the pre-trained model and learn new modules for data-efficient updates, such as the prompt learning methods VideoPrompt \cite{A5_eccv22} and VL Prompting \cite{vificlip_cvpr23}. By identifying and fixing domain-invariant parameters and only updating a small number of domain-specific parameters without introducing new modules during adaptation, \ourmeos offers a great improvement in adaptation efficiency over both of these approaches. We contrast our model also with prototypical network-based methods that focus only on the transferable part, learning the transferable knowledge via meta-learning and directly transferring a metric model for the recognition of new actions~\cite{wang2021few,zhu2021few,fu2020depth,perrett2021temporal} 


In this paper, we propose a method, \ourmeo, that aims to learn domain-invariant features and optimizes the latter efficiently when faced with novel data. 
The recent minimal change principle \cite{causal_icml22} states when adapting a model to new data, one should make the smallest necessary changes to accommodate the new data. 
Taking inspiration from this principle, our key assumption is that domain information can be well estimated and temporal dynamic models are transferable. 
This assumption is inspired by the fact that the laws of physics remain constant across time, and therefore across all frames in any real-world video, and yet these laws are not directly observable. Any knowledge of the laws of physics learned by the pre-trained model should be transferable. We hypothesize that this is a general phenomenon, where temporal invariance in the relations between latent variables yields greater transferability. 

Figure~\ref{fig:toy} exemplifies an example of our assumption by comparing ``Moving Sth.'' to ``Dropping Sth.''. 
For both actions, the movement of a hand leads to the movement of an object. In each frame, we observe the position and orientation of the hand and the object, while their velocity and angle of motion are latent. Newton's laws of motion determine the acceleration (change in velocity) due to gravity in free fall, how the motion of the hand influences the motion of the object, etc. These relations are temporally invariant and can be transferred. Simultaneously, the two samples differ in aspects like the angle of motion of the hand. These aspects can change across the frames from a single sample and hence we expect them to be updated.

Specifically, we employ the temporal causal representation learning~\cite{causal_neurips22,chen2024caring} as the first unsupervised stage. Conversely, we assume that domain information is estimable and therefore does not require domain labels, making it a fully unsupervised learning process. 
Furthermore, we design the representation learning model with four modules: temporal dynamics generation, temporal dynamics transitions, visual encoder, domain estimator, and classifier. Classifier is learned after representation learning stage using labeled action classes while fixing the representation models.  
In the representation learning stage, we use the same temporal dynamic models for different 
domain and visual embeddings, by assuming that 
temporal dynamics are transferable.
During adaption with few-shot examples, we fix the invariant temporal dynamics generation and transitions modules, and update the image encoder, domain estimator, and classifier.
\bfsection{Contributions} 
1) We find that the temporal dynamics are transferable, and thus propose \ourmeo, a new framework with causal representation learning as a preliminary stage to distinguish invariant temporal dynamics and other variant knowledge. 
2) We demonstrate the superior accuracy of \ourmeos over existing models across five standard benchmark datasets and validate our modeling assumptions through comprehensive ablation experiments. Our positive results are consistent with our central assumption that domain-invariance of the learned temporal dynamics suggests transferability to novel contexts.


\section{Related Work}

\bfsection{Few-shot action recognition}
%
%
%
Existing work on few-shot learning uses various approaches to deal with the temporal nature of action recognition~\cite{maml_icml17}. For instance, STRM \cite{few_cvpr22_1} introduces a spatio-temporal enrichment module to look at visual and temporal context at the patch and frame level. HyRSM \cite{few_cvpr22_2} uses a hybrid relation model to learn relations within and across videos in a given few-shot episode.
However, recent work has shown that these methods do not generalize well when the base and novel data have significant distribution disparities \cite{xdomain_cviu23,xdomain_iccv23}.

%
The recent success of cross-modal vision-language learning has inspired works like ActionCLIP \cite{ActionCLIP_arxiv21}, XCLIP is presented by the authors of  \cite{xclip_eccv22}, VideoPrompt \cite{A5_eccv22}, as well ViFi-CLIP is introduced in \cite{vificlip_cvpr23}. Most of these methods apply transfer learning by adopting the pre-trained CLIP and adapting it for few-shot action recognition tasks with popular tuning strategies. For example, VL prompting fixes the backbone of CLIP and tunes the additional visual prompt \cite{vpt_eccv22} to adapt to the novel data. Nevertheless, if the CLIP model is pre-trained on a base dataset that is too dissimilar from the novel data, the issues from distribution disparity persist.
Unlike previous work, our approach to few-shot action recognition leverages intuition from recent advancements in causal representation learning, providing an advantage for handling distribution disparities.

\bfsection{Domain-invariant feature learning}
Domain generalization (DG), or out-of-distribution generalization, aims to enable domain adaptation when the base and novel data are not \textit{i.i.d} and none of the novel data is available when training on the base data. Approaches for achieving DG include meta-learning~\cite{li2018learning}, data augmentation~\cite{shorten2019survey}, and domain-invariant feature learning~\cite{matsuura2020domain,lu2022domain}. Domain-invariant feature learning naturally seeks to learn transferable knowledge that generalizes across domains but to improve the efficacy of few-shot learning we must also work with non-transferable knowledge. 

The way \ourmeos distinguishes the transferable knowledge from the non-transferable knowledge is based on recent advances in causal representation learning on temporal data \cite{causal_iclr22,factoredrl_neurips22,causal_neurips22}. Those works assume the disentanglement of the known auxiliary domain index from the data generation process. In contrast, without known values for the auxiliary variable, we propose to estimate it instead. 

\begin{figure}
\centering
  \includegraphics[width=\linewidth]{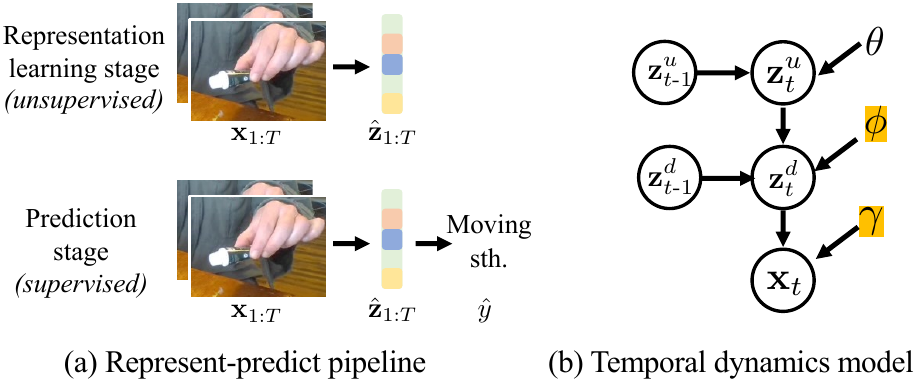}
  \caption{Represent-predict pipeline. (a) We first learn feature representation unsupervisedly for video frames and then supervisedly train the action prediction model. (b) In the ``Represent" stage, we design a temporal dynamics model where parameters (\colorbox{bananayellow}{$\phi$}, \colorbox{bananayellow}{$\gamma$}) are domain-invariant.}
  \vspace{-0.4cm}
  \label{fig:twostage}
\end{figure}

\section{Methodology}
\label{gen_inst}

\subsection{Few-shot Learning Framework Overview}
\bfsection{Two-phase model training}
A typical few-shot setting has two phases of model training on two sets of data: base and novel. 
The phase 1 training is on the base data $\mathcal{D}$, where the video action labels are from the set $\mathcal{C}_{base}$. The novel data consists of two parts, the support set $\mathcal{S}$ for updating the model and the query set $\mathcal{Q}$ for inference. Notably, there only exist limited samples for $\mathcal{S}$, and the video action labels are from the set $\mathcal{C}_{novel}$ that is disjoint from $\mathcal{C}_{base}$.
The phase 2 training updates the model on the support set $\mathcal{S}$, aiming to improve the inference accuracy for novel classes in $\mathcal{Q}$.

\bfsection{Represent-predict pipeline}
For each video clip with $T$ frames, let $\rvx_{1:T}$ be the frame images and $y$ be the action label.
During each phase of model training, we follow the popular approach~\cite{videomae_neurips22} to divide the pipeline into two stages (Fig.~\ref{fig:twostage} (a)).
During the representation learning stage, we train the model to extract feature $\hat{\rvz}_{1:T}$ from frame images $\rvx_{1:T}$ in an unsupervised manner.
During the prediction stage, we fix the feature extractor and train the classifier to predict the action label $\hat{y}$.
In this paper, we apply this framework and learn the causal representation in the first stage to better distinguish the invariant causal dynamics and other variant knowledge. 

\begin{figure*}{}
\centering
  \includegraphics[width=\linewidth]
  {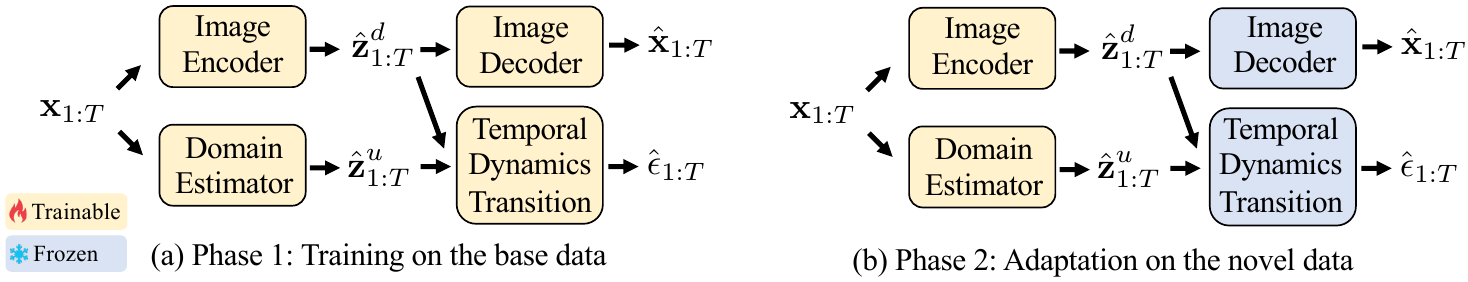}
   \caption{Our \ourmeo\HS model for the representation learning stage. 
  (a) During phase 1 on the base data, we train the whole causal temporal dynamics model. 
  (b) During phase 2 on the novel data, we fine-tune both the image encoder and the domain estimator while freezing the image encoder and the temporal dynamicsOur \ourmeo\HS model for the representation learning stage. 
  (a) During phase 1 on the base data, we train the whole causal temporal dynamics model. 
  (b) During phase 2 on the novel data, we fine-tune both the image encoder and the domain estimator while freezing the image encoder and the temporal dynamics transition module trained during phase 1.
   transition module trained during phase 1.
  }
  \label{fig:structure}
\end{figure*}

\subsection{Causal Temporal Dynamics Model}
Inspired by previous works on temporal causal modeling, we design a generative model based on the variational autoencoder (VAE) in the representation learning stage.

\bfsection{Data generation process}
Each observation $\mathbf{x}_t $ is generated from a nonlinear mixing function $\mathbf{g}$ that maps latent variables $\mathbf{z}_{t}^d $ to $\mathbf{x}_t$, where $\mathbf{z}_{t}^d $ refers to the variables involved in the temporal dynamics.  For every dimension $i \in {1,\dots,m}$ of the latent variable $\mathbf{z}_t^d$, $\mathbf{z}_{t}^d(i)$ is derived from a non-stationary, non-parametric time-delayed causal relation:
\begin{equation}
\label{eq:generation}
\begin{aligned}
         &\underbrace{ \mathbf{x}_t = \mathbf{g}(\mathbf{z}_{t}^d) }_{\text{Generation process}}, \\  &\underbrace{z_{t}^d(i) = 
         {f_i\left(\{z_{t'}^d(j) \vert z_{t'}^d(j) \in \mathbf{Pa}(z_{t}^d(i)) \},z_{t}^u(i) ,\epsilon_{t}(i)  \right)}}_{\text{Non-stationary non-parametric transition}}.
\end{aligned}
\end{equation}
The non-stationary information is captured by the representation $z_{it}^u$ and we assume that this information can be estimated from observed data. Here, we use the superscripts \(d\) and \(u\) to denote dynamic and domain information, respectively.
The validity of this assumption is strong but reasonable since humans can typically estimate the domain information from the observed videos. We also verify the validity of this assumption in the experiments. 

\bfsection{Probabilistic formulation}
Without specifying, our notations follow the TDRL framework~\cite{causal_neurips22}.
As shown in part (b) of Figure~\ref{fig:twostage}, we formulate the temporal dynamics model with $\{\rvx,\rvz^d,\rvz^u\}$. Specifically, for the convenience, we formulate the joint distribution with setting time-lag as 1:
\begin{align}\label{eq:dg}
    p(\rvx_{1:T},&\rvz^d_{1:T},\rvz^u_{1:T}) = p_\gamma(\rvx_1|\rvz^d_1)p_\phi(\rvz^d_1|\rvz^u_1)p(\rvz^u_{1})\nonumber\\
    &\prod_{t=2}^{T}p_\gamma(\rvx_t|\rvz^d_t)\underbrace{p_\phi(\rvz^d_t|\rvz^d_{t-1},\rvz^u_t)}_\text{temporal dynamics}{p(\rvz^u_t|\rvz^u_{t-1})}.
\end{align}
In the VAE framework, one can consider \(\gamma\) as the parameters for the image decoder (generation process) \(p_\gamma(\mathbf{x}_t | \mathbf{z}_t^d)\), and \(\phi\) as the parameters for the temporal dynamics transition \(p_\phi(\rvz^d_t|\rvz^d_{t-1}, \rvz^u_t)\). 
To learn visual representation $ \mathbf{z}_t^d $ and domain information $\mathbf{z}_t^u$
, we also introduce two modules including image encoder $q_{\omega}(\mathbf{z}_t^d|\mathbf{x}_t)$ with parameters $\omega$ and domain estimator $q_{\theta}(\mathbf{z}_t^u| \mathbf{z}_{t-1}^u, \rvx_{t})$ with parameters $\theta$.

\subsection{Network Designs}

\bfsection{Image encoder $q_\omega(\rvz^d_t|\rvx_t)$}
We denote the image encoder using $q_\omega(\hat{\rvz}^d_{t} |\rvx_{t})$. We assume $\rvz^d_t$ is conditionally independent of all $\rvz^d_{t'}$ for $t' \neq t$ conditioned on $\rvx$, and therefore we can decompose the joint probability distribution of the posterior by $q_\omega(\hat{\rvz}^d | \rvx) = \prod\limits_{t=1}^T q_\omega(\hat{\rvz}_{t}^d |\rvx_{t} )$. We choose to approximate $q$ by an isotropic Gaussian characterized by mean $\mu_t$ and covariance $\sigma_t$. To learn the posterior we use an encoder composed of an MLP followed by leaky ReLU activation:
\begin{align}\label{eq:posterior_implement}
    \hat{\rvz}^d_t\sim\mathcal{N}(\mu_t,\sigma_t),~~    
    \mu_t,\sigma_t = \text{LeakyReLU}(\text{MLP}(\rvx_t)).
\end{align}

\bfsection{Domain estimator $q_{\theta}(\mathbf{z}_t^u|\mathbf{z}_{t-1}^u , \rvx_{t})$ } Since we assume that $\rvz^u_t$ captures the domain variance for the transition of $\rvz_t$, $\rvu_t$ needs to be updated when adapting. In our task, we assume that the prior $p(\rvz^u_t)$ is not observed but we can estimate it. Specifically, we build the domain estimator on latent domain learning \citep{adapter_iclr22}. For each domain, we use a set of learnable gating functions, denoted as ${A_1, A_2,\ldots, A_S}$, to obtain $\hat{\rvz}_t^u$:
\begin{align}\label{eq:adapter}
\hat\rvz^u_t = \text{GRU}(\hat\rvz^u_{t-1}, \sum\limits_{\rvs=1}^{S}A_\rvs(\rvx_t)\text{Conv}_\rvs(\rvx_t)).
\end{align}
Here, Conv denotes a $1\times 1$ convolution, enabling linear transformations for $\rvx$. The GRU models any temporal dependencies. We refer the readers to \citep{adapter_iclr22} for more details. 

\bfsection{Temporal Dynamics Transition $p_\phi(\rvz_t^d|\rvz^d_{t-1},\rvz^u_t)$} 
Since the prior distribution of $p(\rvz^d_t|\rvz^d_{t-1},\rvz^u_t)$ is not tractable, we employ a flow network to convert this computation with the distribution of the Gaussian noise $\epsilon_{t} $. 
For $i-$th dimension of the noise vector, we formulate the prior module as $\hat{\epsilon}_{t}(i)=\hat{f}^{-1}_i(\hat{z}_{t}^d(i)|\hat{\rvz}_{t-1}^d,\hat{z}_{t}^u(i) )$. 
Then the prior distribution of the $i$-th dimension of the temporal dynamics, $\hat{\rvz}_{t}^d(i)$, can be computed as 
\begin{align}\label{eq:learnprior}
p_\phi\big(\hat{f_i}^{-1}(\hat{\rvz}_{t}^d(i) | \hat{\rvz}_{t-1}^d, \hat{\rvz}_{t}^u(i))\big)\vert\frac{\partial\hat{f_i}^{-1}}{\partial\hat{\rvz}_{t}^d(i)}\vert.
\end{align}
This flow model is built with the MLP layers.
For a detailed derivation please refer to Appendix~\ref{ap:derive}.

\bfsection{Image decoder $p_\gamma(\rvx_t|\rvz^d_t)$} The image decoder pairs with our image encoder to generate an estimate of the action representations $\hat{\rvx}_t$ from the estimated latent variables $\hat{\rvz}^d_t$. We estimate $p_\gamma(\rvx_t|\rvz^d_t)$ using the decoder, which consists of a stacked MLP followed by leaky ReLU activation: 
\begin{align}\label{eq:decoder}
    \hat{\rvx}_t = \text{LeakyReLU}(\text{MLP}(\hat{\rvz}^d_t)).
\end{align}

\bfsection{Classifier $p_\psi(y|\rvz^d_{1:T})$}
We have two choices of $\psi$. 

The first one is to the end of obtaining the standard one-hot embedding.
We first concatenate $\hat{\rvz}_{1:T}^d$ along the time dimension and apply an MLP model for classification:
\begin{align}
    \hat{y} = \text{MLP}(\text{Concat}(\hat{\rvz}^d_{1:T})).
\end{align}

The alternative is the text embedding layer:
\begin{align}
    \hat{e}=\text{embed}(y)
\end{align}
We implement the text embedding layer using a transformer-based text encoder with eight attention heads, following the architecture used in \cite{vificlip_cvpr23}. 

\subsection{Training Objectives}
\bfsection{Representation learning stage}
\ourmeos first employs the ELBO loss to learn $\gamma,\phi,\theta$,and $\omega$ aforementioned. The ELBO loss combines the reconstruction loss from using our decoder to estimate $\rvx_t$ with the KL-divergence between the posterior and prior:
\begin{align}\label{eq:elbo}
& \mathcal{L}_{\text{ELBO}} =
\underbrace{\frac{1}{T}\sum\limits_{t=1}^{T}\text{BCE}(\rvx_{t},\hat{\rvx}_t)}_{\mathcal{L}_{\text{Recon}}} \nonumber\\ & -  \underbrace{\frac{1}{T}\sum\limits_{t=1}^{T}\beta\mathbb{E}_{\mathbf{\hat \rvz}_{t}^d \sim q_\omega} \big(\log q\left(\mathbf{\hat\rvz}_{t}^d \vert \rvx_{t} \right) - \log p(\mathbf{\hat \rvz}_{t}^d|\hat{\rvz}_{t-1}^d, \rvz_t^u)\big)}_{\mathcal{L}_{\text{KLD}}}.
\end{align}
Here, $\mathcal{L}_{\text{Recon}}$ measures the discrepancy between $\rvx_t$ and $\hat{\rvx}_t$ using binary cross-entropy (BCE), and $\beta$ is the hyperparameter to balance the two losses.

\bfsection{Prediction stage}
Depending on the choice of the classifier, we use the corresponding loss function. 
The first choice is to use the CE loss using a one-hot encoding of $y$, 
\begin{align}\label{eq:ce_loss}
\mathcal{L}^{CE}_\text{cls}=-\mathbb{E}_{ \hat{y}}\big(\text{one-hot}(y)\cdot\log(\text{softmax}(\hat{y}))\big).
\end{align}

Another choice is the NCE loss \citep{nce_2010}.
For the $k^\text{th}$ video sequence, let $m$ be the indices of the action labels.
We use the temporal pooling over $\hat{\mathbf{z}}^d$ obtain $\hat{\mathbf{z}}^d_{1:T,k}$.

\begin{align}\label{eq:contrastive_loss}
\mathcal{L}^{NCE}_\text{cls}=-\sum\limits_{k}\log\frac{\text{exp}(\text{sim}(\hat{\mathbf{z}}^d_{1:T,k},\hat{e}_k)/\tau)}{\sum\limits_{m}\text{exp}(\text{sim}(\hat{\mathbf{z}}^d_{1:T,k},\hat{e}_m)/\tau)},
\end{align}
where $\text{sim}(\cdot, \cdot)$ is the cosine similarity between the text embedding of the action label and our learned $\hat{\mathbf{z}}^d_{1:T,k}$ and $\tau$ is the temperature parameter. 
In practice, we use Eq.~\ref{eq:ce_loss} or Eq.~\ref{eq:contrastive_loss} \cite{vificlip_cvpr23} for a fair comparison with previous methods, denoted as \ourmeo$_{CE}$ and \ourmeo$_{NCE}$ respectively. 
%

\subsection{Training, Adapting and Inference}

The overall framework is illustrated in Figure~\ref{fig:structure}. The model training involves two phases: in Phase 1, we train all modules with base data, while in Phase 2, we only adapt a part of modules with novel few-shot data.

\bfsection{Phase 1: Training on the base data $\mathcal{D}$}
For the basic training, we first learn the causal temporal dynamics in an unsupervised way. Specifically, we jointly optimize the image encoder ($\omega$), domain estimator ($\theta$), dynamics transition ($\phi$), and the image decoder ($\gamma$) with the ELBO loss in equation~\ref{eq:elbo}. Then, in the next stage, we fix the whole model and learn the classifier $\psi$ with either CE or NCE losses in Eq.~\ref{eq:ce_loss} and~\ref{eq:contrastive_loss}.

\bfsection{Phase 2: Adaptation on the novel data support set $\mathcal{S}$}
Given the assumption that temporal dynamics are invariant on both base and novel data, we keep the learned temporal dynamics model unchanged in this phase.
Specifically, in the representation stage for adaptation of the novel data, we only update the image encoder ($\omega$) and domain encoder ($\theta$). Then, we update the prediction model parameter ($\psi$) with a few shot labels. 

\bfsection{Inference on the novel data query set $\mathcal{Q}$}
To perform inference, we sample $\hat{\rvz}^d$ according to Eq.~\ref{eq:posterior_implement} using our image encoder, and we either choose the maximum value (highest probability) from the prediction $\hat{y}$ or choose the label whose text embedding maximizes cosine similarity.

\section{Experiments}\label{sec:exp} 

\subsection{Experimental Setup}


We carry out two types of few-shot learning experiments; all-way-k-shot and 5-way-k-shot. \cite{A5_eccv22} proposes the setting of all-way-k-shot: we try to classify all action classes in the class for the novel dataset $(\mathcal{C}_{novel})$, while in 5-way-k-shot learning we only try to estimate 5 label classes at a time in a series of trials. The number of shots (k) refers to the number of training samples in $\mathcal{S}$ available for each action label. 
Once we partition our data into base set $\mathcal{D}$ and novel set $\mathcal{S} \cup \mathcal{Q}$, the number $k$ determines how many samples for each action class we choose for $\mathcal{S}$ and the rest are used for the query set $\mathcal{Q}$.

\bfsection{Datasets}
In this work, we conduct experiments on five datasets: 
1. Something-Something v2 (SSv2) is a dataset containing 174 action categories of common human-object interactions;
2. Something-Else (Sth-Else) exploits the compositional structure of SSv2, where a combination of a verb and a noun defines an action; 
3. HMDB-51 contains 7k videos of 51 categories; 
4. UCF-101 covers 13k videos spanning 101 categories; 
5. Kinetics covers around 230k 10-second video clips sourced from YouTube. 
For the experiments that perform training and testing on the Sth-Else dataset, we use the official split of data~\cite{sthelse_cvpr20,sthelse_cvpr22,sthelse_neurips22}.
Similarly, for experiments using other datasets, we split the data into $\mathcal{D}$, $\mathcal{S}$, and $\mathcal{Q}$ following the prior work we use as benchmarks, as described late. The other datasets we use for novel data are SSv2~\cite{sthdata_iccv17}, SSv2-small~\cite{sssmall_eccv18}, HMDB-51~\cite{hmdb_iccv21}, and UCF-101~\cite{ucf_arxiv12}.
\bfsection{Evaluation metrics}
In all experiments, we compare the Top-1 accuracy, i.e., the maximum accuracy on any action class, of \ourmeos against leading benchmarks for few-shot action recognition. The results of the top-performing model are given in \textbf{bold}, and the second-best are \underline{underlined}. 

\bfsection{Implementation details}
In each experiment, there is a backbone used to extract the action representations $\rvx_t$ used for training, and in all experiments, the backbone is either ResNet-50~\cite{resnet_cvpr16} or ViT-B/16~\cite{clip_icml21} trained on the base data set $\mathcal{D}$, where the choice of backbone matches what was used in the benchmark experiments. 
We use the AdamW optimizer \cite{adamw_iclr19} and cosine annealing to train our network with a learning rate initialized at 0.002 and weight decay of 10$^{-2}$.
For all video sequences we use $T=16$ uniformly selected frames. To compute the ELBO loss, we choose $\beta=0.02$ to balance the reconstruction loss and KL-divergence. Also, we set $\tau=0.07$ for the NCE loss. Regarding the hyperparameters, we set $d=12$ in Eq.~\ref{eq:generation}, and $S=35$ in Eq.~\ref{eq:adapter}. The hyperparameter analysis is reported in Table~\ref{tab:ablation_higherorder}.
For a detailed summary of our network architectures see Table~\ref{tab:architecture} in the appendix.
Our models are implemented using PyTorch, and experiments are conducted on four Nvidia GeForce 2080Ti graphics cards.

\subsection{Benchmark Results}\label{sec:benchmark}

In the first part of our experiment, we compare the results from \ourmeos against state-of-the-art few-shot action recognition methods. 
To obtain fair comparisons, we compare \ourmeos with the leading approaches using identical backbones and classification loss on each dataset.

\begin{table}[t]
\centering
 \caption{Benchmark results on the Sth-Else dataset with the same ViT-B/16 backbone but two different loss functions.
 }
 \scalebox{0.9}{
\begin{tabular}{l|l|c|cc}
\hline
{}
& \multirow{2}{*}{Loss}
&\multicolumn{2}{c}{Sth-Else}
\\ \cline{3-4}
&
&{5-shot}
&{10-shot}
\\ \hline 
ORViT~\cite{sthelse_cvpr22} &  & 33.3 & 40.2 \\
SViT~\cite{sthelse_neurips22} & CE & \underline{34.4} & \underline{42.6}\\
\bf{\ourmeo$_{CE}$ (ours)} &  & \bf{37.6} & \bf{44.0} \\
\hdashline
ViFi-CLIP~\cite{vificlip_cvpr23} &  & 44.5 & 54.0 \\
VL Prompting~\cite{vificlip_cvpr23} & NCE & \underline{44.9} & \underline{58.2} \\
\bf{\ourmeo$_{NCE}$  (ours)} &  & \bf{48.5} & \bf{63.9} \\
\hline
\end{tabular}
}
 \label{tab:sth-else}
 \vspace{-0.3cm}
\end{table}

\bfsection{Sth-Else Experiments}
Table ~\ref{tab:sth-else} shows two experiments against benchmarks for all-way-k-shot learning using the ViT-B/16 backbone for the Sth-Else dataset.
We compare \ourmeo$_{CE}$ to leading benchmarks on Sth-Else that use cross-entropy loss, ORViT and SViT~\cite{sthelse_cvpr22,sthelse_neurips22}.
We also compare \ourmeo$_{NCE}$ to the state-of-the-art methods that employ contrastive learning, ViFi-CLIP, and VL-Prompting~\cite{vificlip_cvpr23}.

It is evident that \ourmeo$_{NCE}$ and \ourmeo$_{CE}$ had the highest accuracy in both experiments. \ourmeos outperforms all four leading benchmarks by amounts ranging from 1.4 to 5.7 percentage points. For this dataset we see the greatest improvement for $k=5$, improving from 34.4 to 37.6 and from 44.9 to 48.5 over the leading benchmarks, and with improvements from 58.2 to 63.9 for $k=10$.

\begin{figure}[t]
\centering
\includegraphics[width=\linewidth]
{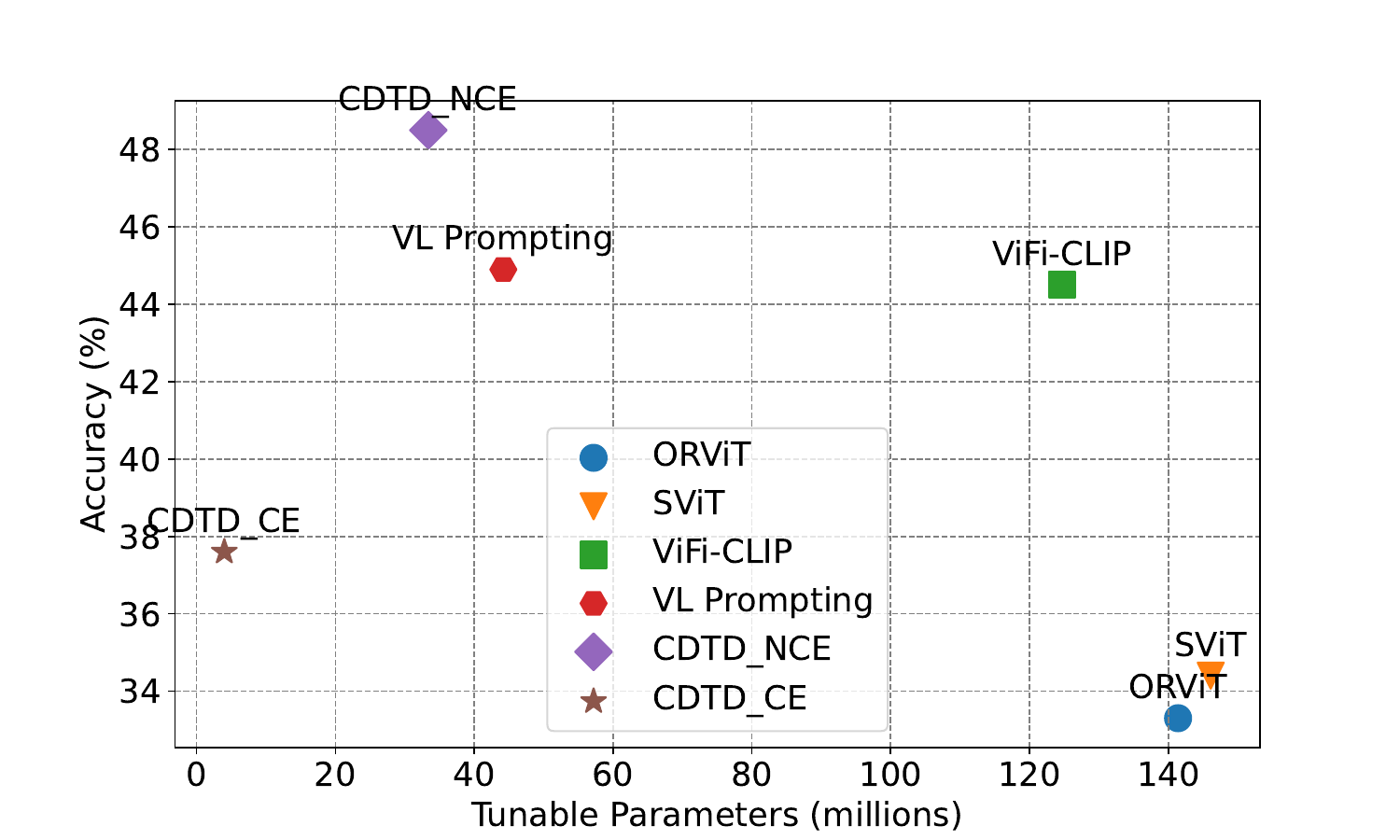}
\caption{The comparison of the parameters that need to be updated during adapting on \textbf{Sth-Else dataset}.}
\label{fig:params}
 \vspace{-0.3cm}
\end{figure}

Figure~\ref{fig:params} further shows the number of parameters updated during transfer.
Our \ourmeos necessitates tuning the fewest parameters among all the approaches when adapting from the base data to the novel data, bringing great parameter efficiency. Notably, ViFi-CLIP, ORViT, and SViT require an order of magnitude more parameters to be tuned for adaptation to novel data compared to \ourmeo. Similarly, the VPT \cite{vpt_eccv22} of CLIP  needed for VL Prompting results in more than \textbf{40 million} parameters needing updating.

\begin{figure*}
\begin{center}
\includegraphics[width=\linewidth]{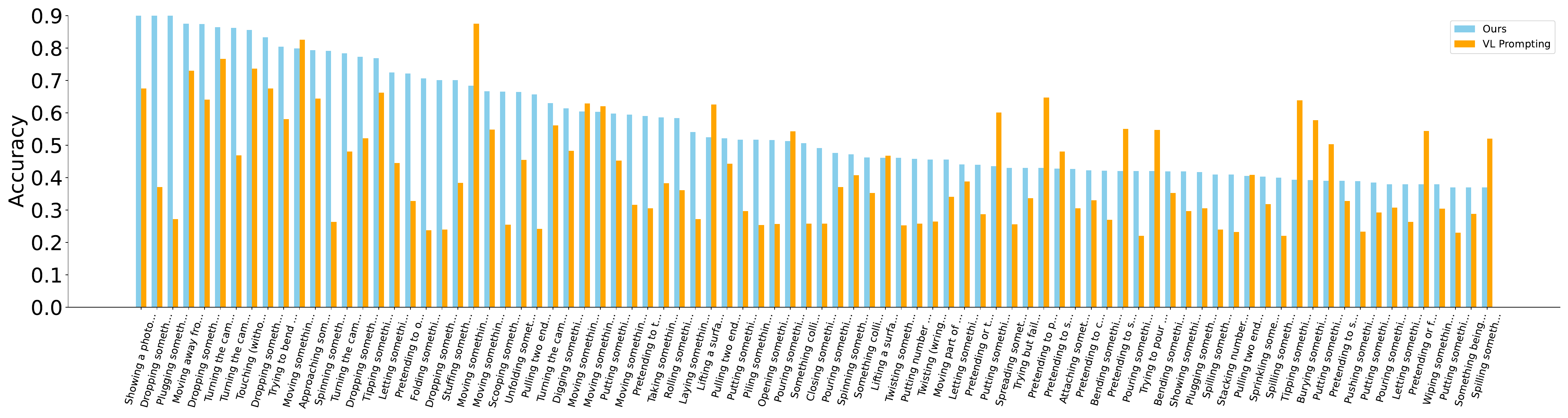}
\end{center}
\caption{Comparing performance of \ourmeo$_{NCE}$ (blue) against VL-Prompting (orange) across all action classes on the Sth-Else dataset.}
\label{fig:visual}
    
\end{figure*}
Fig.~\ref{fig:visual} shows that \ourmeos outperforms VL-Prompting on the majority of action classes. Notably, \ourmeos exceeds the VL prompt in 72 out of 86 action classes, such as “plugging something into something” and “spinning so it continues spinning.” We also note significant improvements in other categories, including “approaching something with something” and “pulling two ends of something but nothing happens”.
However, \ourmeos does have some limitations. Specifically, VL-Prompting surpasses our method in label classes like ``burying something in something'' and ``lifting a surface with something on it''. 

\bfsection{All-way-$k$-shot Experiments}
Table~\ref{tab:allway_p2} show the experimental results using the ViT-B/16 backbone comparing \ourmeo$_{NCE}$ to leading benchmarks for all-way-k-shot learning for $k \in \{2,4,8,16\}$.
For all-way-k-shot learning, our benchmarks are: ActionCLIP \cite{ActionCLIP_arxiv21}, XCLIP \cite{xclip_eccv22}, VideoPrompt \cite{A5_eccv22}, VL Prompting and ViFi-CLIP \cite{vificlip_cvpr23}, VicTR \cite{victr_arxiv23} and VideoMAE \cite{videomae_neurips22}.

\begin{table*}[h]
\centering
\caption{Benchmark results for the all-way-k-shot learning setting. All models use ViT-B/16 backbone and the NCE loss.}
\label{tab:allway_p2}
\scalebox{0.8}{
\begin{tabular}{l|cccc|cccc|cccc}
\hline
{}
&\multicolumn{4}{c}{SSv2}
&\multicolumn{4}{c}{HMDB-51}
&\multicolumn{4}{c}{UCF-101}
\\ \hline
{}

&{2-shot}
&{4-shot}
&{8-shot}
&{16-shot}
&{2-shot}
&{4-shot}
&{8-shot}
&{16-shot}
&{2-shot}
&{4-shot}
&{8-shot}
&{16-shot}
\\ \hline 
XCLIP\cite{xclip_eccv22}  & 3.9 & 4.5 & 6.8 & 10.0 & 53.0 & 57.3 & 62.8 & 64.0 & 70.6 & 71.5 & 73.0 & 91.4  \\
ActionCLIP\cite{ActionCLIP_arxiv21} & 4.1 & 5.8 & 8.4 & 11.1  & 47.5 & 57.9 & 57.3 & 59.1 & 70.6 & 71.5 & 73.0 & 91.4  \\
VicTR\cite{victr_arxiv23} &  4.2 & 6.1 & 7.9 & 10.4 & 60.0 & 63.2 & 66.6 & 70.7   & 87.7 & 92.3 & 93.6 & 95.8 \\
VideoPrompt\cite{A5_eccv22} & 4.4 & 5.1 & 6.1 & 9.7 & 39.7 & 50.7 & 56.0 & 62.4 & 71.4 & 79.9 & 85.7 & 89.9  \\
ViFi-CLIP\cite{vificlip_cvpr23} & 6.2 & 7.4 & 8.5 & 12.4 & 57.2 & 62.7 & 64.5 & 66.8 & 80.7 & 85.1 & 90.0 & 92.7 \\
VL Prompting\cite{vificlip_cvpr23} & 6.7 & 7.9 & 10.2 & 13.5 & 63.0 & 65.1 & 69.6 & 72.0 & \bf{91.0} & 93.7 & \underline{95.0} & 96.4 \\
VideoMAE \cite{videomae_neurips22} & \underline{8.2} & \underline{10.0} & \underline{15.1} & \underline{18.2} & \underline{63.7} & \underline{69.4} & \underline{70.9} & \underline{75.3} & \bf{91.0} & \underline{94.1} & 94.8 & \underline{97.7} \\
\bf{\ourmeo$_{NCE}$ (ours)} & \bf{9.5} & \bf{11.6} & \bf{14.8} & \bf{19.5}  & \bf{65.8} & \bf{70.2} & \bf{72.5} & \bf{77.9}  & \underline{90.6} & \bf{94.7} & \bf{96.2} & \bf{98.5}\\
\hline
\end{tabular}
}
\end{table*}

We follow~\cite{ActionCLIP_arxiv21,xclip_eccv22,A5_eccv22,vificlip_cvpr23} in using Kinetics-400 (K-400) from~\cite{kinectis_cvpr17} as the base dataset $\mathcal{D}$ and repeat the experiment using novel data from the SSV2, HMDB-51, and UCF-101 datasets. 
We can observe that \ourmeo$_{NCE}$ had the highest accuracy in 11 out of 12 of these experiments and had the second highest accuracy in the remaining experiment trailing by only 0.4.

\begin{table*}[h]
\centering
\small 
\caption{Benchmark results for the 5-way-k-shot learning setting. All models use the ResNet-50 backbone and the CE loss.}
\label{tab:nway_p1_main}
\scalebox{0.9}{
\begin{tabular}{l|cccc|cccc}
\hline
{}
&\multicolumn{4}{c}{1-shot}
&\multicolumn{4}{c}{5-shot}
\\ \hline
{}
&{UCF-101}
&{HMDB-51}
&{SSv2}
&{SSv2-small}
&{UCF-101}
&{HMDB-51}
&{SSv2}
&{SSv2-small}
\\ \hline 
OTAM\cite{few_cvpr20} & 50.2 & 34.4 & 24.0 & 22.4 & 61.7 & 41.5 & 27.1 & 25.8\\
TRX\cite{few_cvpr21} & 47.1 & 32.0 & 23.2 & 22.9 & 66.7 & 43.9 & 27.9 & 26.0\\
STRM\cite{few_cvpr22_1} & 49.2 & 33.0 & 23.6 & 22.8 & 67.0 & 45.2 & 28.7 & 26.4\\
DYDIS\cite{xdomain_neurips21} & 63.4 & 35.2 & 25.3 & 24.8 & 77.5 & 50.8 & 29.3 & 27.2\\
STARTUP\cite{xdomain_iclr21} & \underline{65.4} & 35.5 & 25.1 & 25.0 & 79.5 & 50.4 & 31.3 & 28.7\\
SEEN\cite{xdomain_cviu23} & 64.8 & \underline{35.7} & \underline{26.1} & \underline{25.3} & \bf{79.8} & \underline{51.1} & \underline{34.4} & \underline{29.3}\\
\bf{\ourmeo$_{CE}$ (ours)} & \bf{66.0} & \bf{37.6} & \bf{31.4} & \bf{28.5} & \underline{79.7} & \bf{54.4} & \bf{37.0} & \bf{32.8}\\
\hline
\end{tabular}
}
\end{table*}
 
\bfsection{5-way-$k$-shot Experiments} 
Table~\ref{tab:nway_p1_main} shows the results of experiments comparing \ourmeo$_{CE}$ to leading benchmarks for 5-way-k-shot learning. All models are trained on K-400 as the base data using the ResNet-50 backbone. In each trial, we select 5 action classes at random and update our model using only $k$ samples for each of those classes to form $\mathcal{S}$, and the remaining novel data compose $\mathcal{Q}$. 
To ensure the statistical significance, we conduct 1000 trials with random samplings for selecting the action classes in each trial by following \cite{xdomain_cviu23}, and report the mean accuracy as the final result.
Our leading benchmarks are: OTAM \cite{few_cvpr20}, TRX \cite{few_cvpr21}, STRM \cite{few_cvpr22_1}, DYDIS \cite{xdomain_neurips21}, STARTUP \cite{xdomain_iclr21}, and  SEEN \cite{xdomain_cviu23}.
\ourmeo$_{CE}$ has the highest accuracy in 7 out of 8 experiments, by a margin ranging from 0.6 to 5.3, and has the second highest accuracy on the remaining experiment, trailing by only 0.1. 


\subsection{Ablation Studies}\label{sec:ablation}

\begin{table}[h]
\centering
 \caption{Ablation study results on the effect of each component on the Sth-Else dataset. }
\begin{tabular}{l|cc}
\hline
&{5-shot}
&{10-shot}
\\ \hline
w/o temporal dynamic transition & 41.0 & 44.3 \\
w/o domain encoder & 42.8 & 50.6 \\
w/o temporal modeling & 45.1 & 54.4 \\
\ourmeo$_{NCE}$-FT &  44.8 & 58.3\\
\ourmeo$_{NCE}$-UG &  46.0 & \underline{61.3}\\
\ourmeo$_{NCE}$-UT &  \underline{46.2} & 60.8 \\
\bf{\ourmeo$_{NCE}$} & \bf{48.5} & \bf{63.9} \\
\hline
\end{tabular}
 \label{tab:ablation}
\end{table}

\bfsection{Components of \ourmeo}
In this section, we demonstrate the contribution of each component of our \ourmeos model. We proceeded by comparing our \ourmeos model to four simpler baselines: w/o temporal dynamic transition, w/o domain encoder, w/o temporal modeling, and the full model with all components fine-tuned \ourmeo-FT. The model w/o temporal dynamic transition removes the temporal dynamic transition and domain encoder. The posterior is regularized by KL-divergence with the standard normal distribution. The model w/o temporal modeling assumes $\rvz_{t-l} = \rvz_{t}$ for all $t$. Per time step, the posterior in w/o temporal baseline regularizes the prior by KL-divergence independently from other time steps. The model w/o domain encoder removes the domain encoder module so only the image encoder $\omega$ and classifier $\phi$ are updated during adaptation. For \ourmeo-FT, we also finetune the temporal dynamic transition and image decoder rather than fixing them during adaptation. \ourmeo-UG updates the temporal dynamic transition module during adaptation, while the temporal dynamic generation module remains fixed. \ourmeo-UT updates all modules during the adaptation process.

We summarize the results of our ablation study in Table~\ref{tab:ablation}. First, notice that the accuracy of 45.1 by the w/o temporal modeling is already better than the 44.5 and 44.9 achieved by ViFi-CLIP and VL-prompting, respectively~\cite{vificlip_cvpr23} (Table~\ref{tab:sth-else}). This demonstrates the superior transportability of our method. Moreover, \ourmeos performs even better because it captures the temporal relations of the latent causal variables, and $\vtheta$ and $\omega$ help capture the distribution shift across data sets. The superiority of \ourmeos over both w/o causal and w/o temporal demonstrates the advantage of modeling the action generation process.

\bfsection{Hyperparamter sensitivity}
\ourmeos has three hyperparameters: the numbers of latent variables $d$ in Eq.~\ref{eq:generation}, and the number of latent domains $S$ in Eq.~\ref{eq:adapter}. We also extend the time lags by choosing $l=1$ to $l=3$ in Eq.~\ref{eq:dg}
We report the results varying these hyperparameters in Table~\ref{tab:ablation_higherorder} using \ourmeo$_{NCE}$ on the Sth-Else dataset. We vary one variable at a time, keeping the others constant. 

\begin{table}[h]
\centering
\caption{Ablation study results on the hyperparameter sensitivity on the Sth-Else dataset.}
{
\begin{tabular}{l|cc}
\hline
{}
&{5-shot}
&{10-shot}
\\ \hline 
$d=4$ & 44.2 & 49.0  \\
$d=8$ & 46.0 & 55.7  \\
$d=12$ & \underline{48.5} & \underline{63.9}  \\ 
$d=16$ & \bf{48.9} & \bf{65.1}  \\
\hdashline
$l=1$ & \underline{48.5} & 63.9  \\
$l=2$ & \underline{48.5} & \underline{64.0}   \\
$l=3$ & \bf{48.8} & \bf{64.6}   \\
\hdashline
$S=10$ & 42.5 & 51.1  \\
$S=20$ & 46.4 & 59.8  \\
$S=35$ & \underline{48.5} & \underline{63.9}  \\ 
$S=50$ & \bf{49.1} & \bf{65.0}  \\
\hline
\end{tabular}
}
 \label{tab:ablation_higherorder}
\end{table}

Based on our ablation experiments in Table~\ref{tab:ablation_higherorder} we fix $d=12$, $l=1$, and $S=35$ in all of our experiments unless otherwise stated.
\ourmeo$_{NCE}$ performs better with bigger $d$, verifying that the latent causal variables facilitate action representation.
We choose $d = 12$ because beyond this point the performance gain becomes marginal and the computational cost spikes.
If we allow the parents of each $\rvz$ to be further back in time $(l>1)$, allowing causal relationships between non-consecutive time steps, the change in accuracy is negligible. Therefore, we choose that the parents of each $\rvz$ are only in the previous time step $(l=1)$.
Finally, a consistent improvement is observed when increasing $S$ from 10 to 50. However, considering the increased computational cost associated with the linear transformations and $1\times 1$ convolutions involved in Eq.~\ref{eq:adapter}, we have opted for $S=35$ as the standard across our experiments.

\begin{figure}[t]
\captionsetup[subfloat]{labelformat=simple}
\centering
\subfloat[\label{fig:motivating_a}]{
\includegraphics[width=0.5\linewidth]
{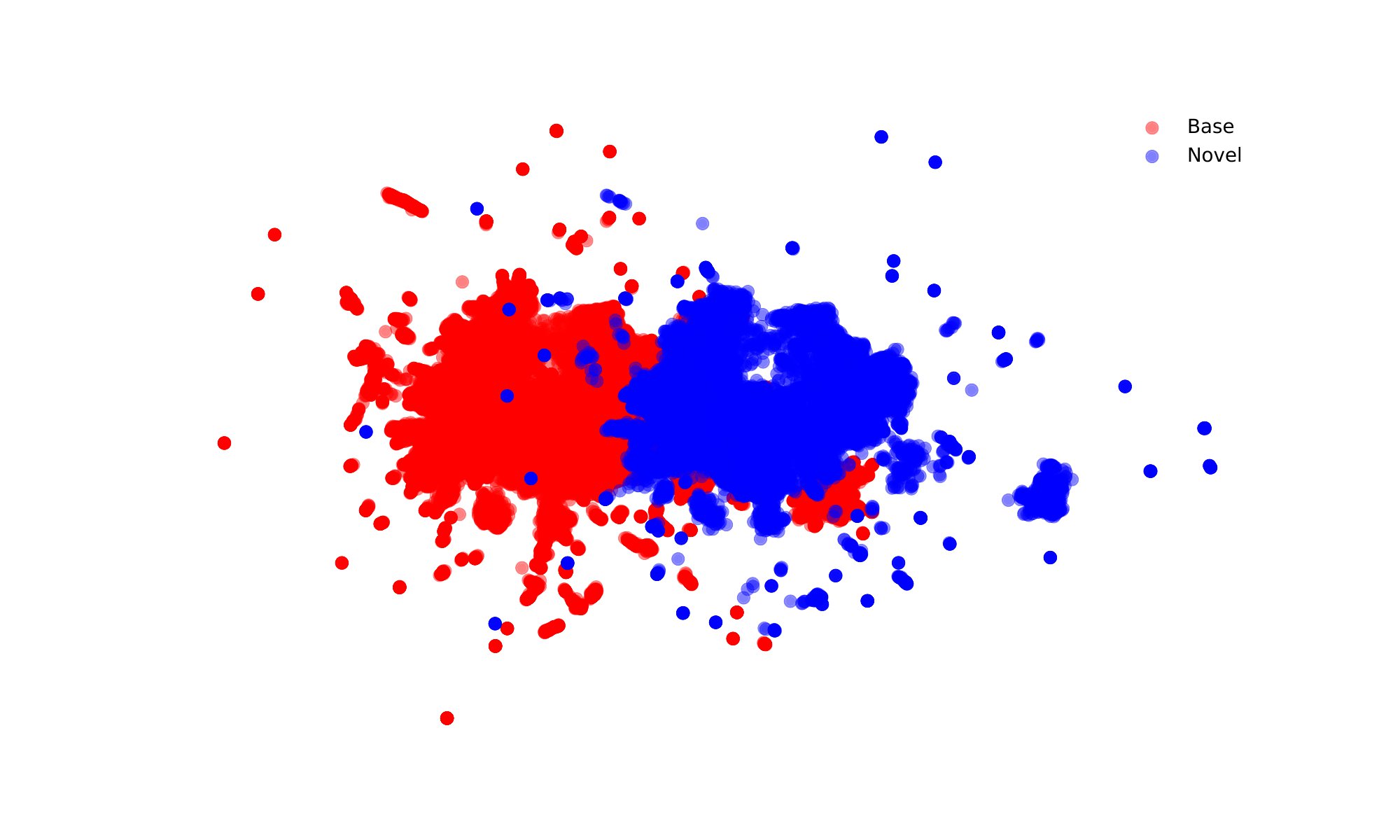}}
\subfloat[\label{fig:motivating_c}]{
\includegraphics[width=0.5\linewidth]
{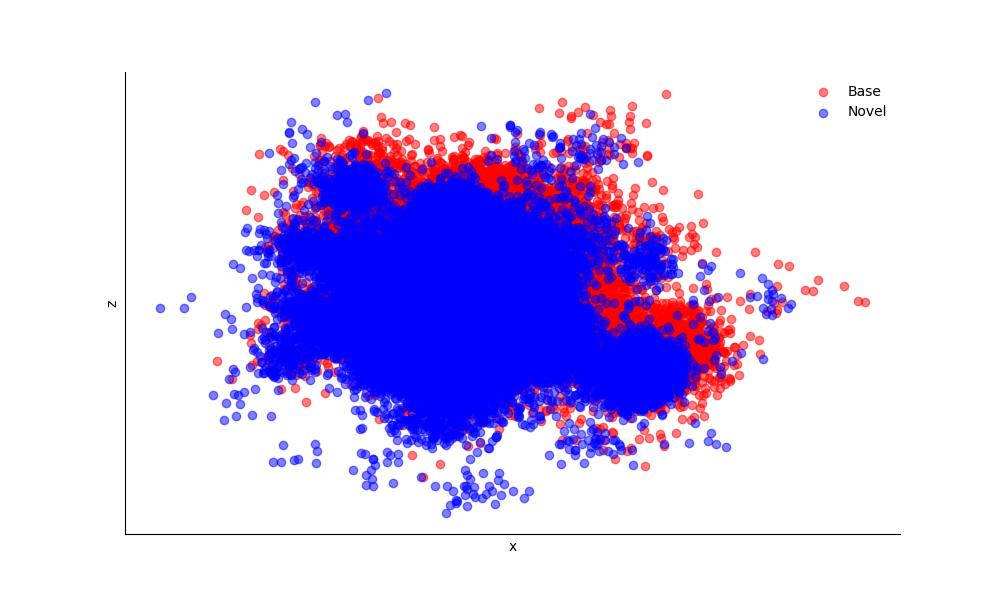}}\\
\subfloat[\label{fig:motivating_d}]{
\includegraphics[width=0.5\linewidth]
{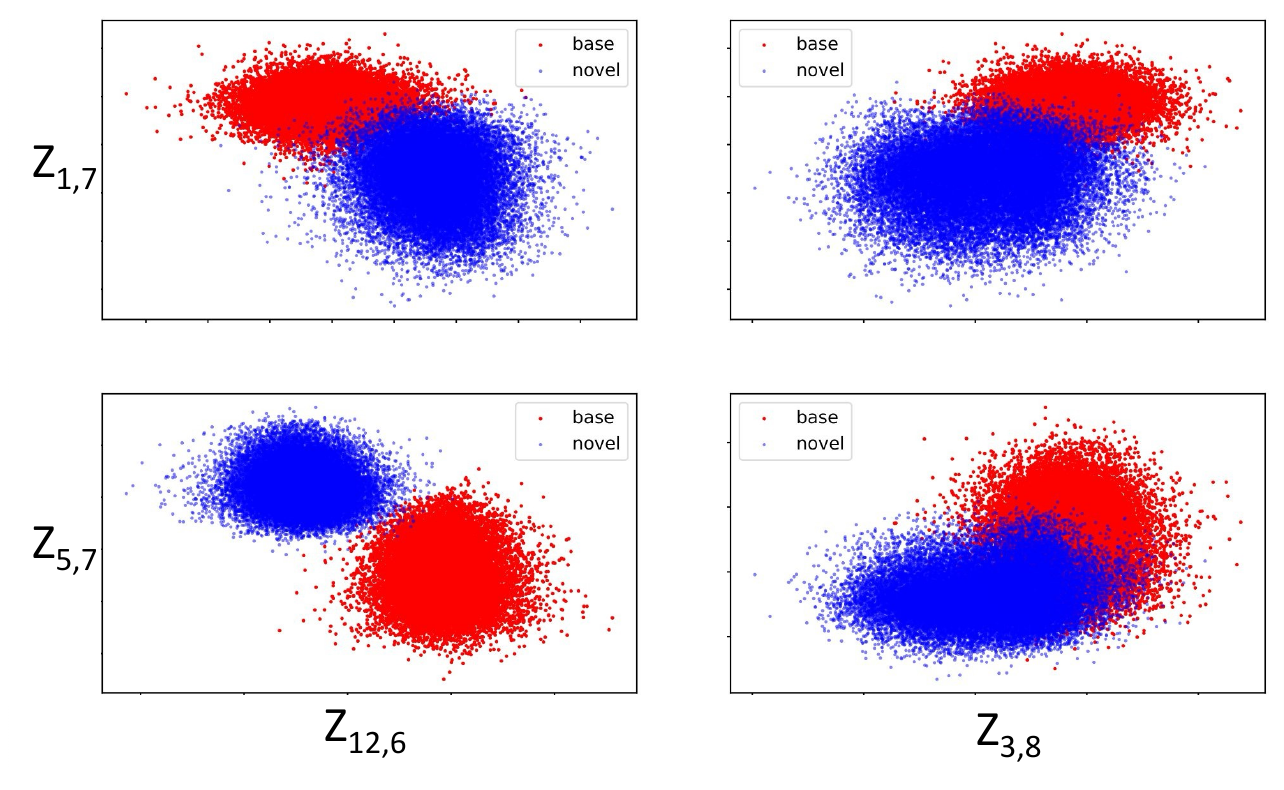}}
\subfloat[\label{fig:motivating_b}]{
\includegraphics[width=0.5\linewidth]
{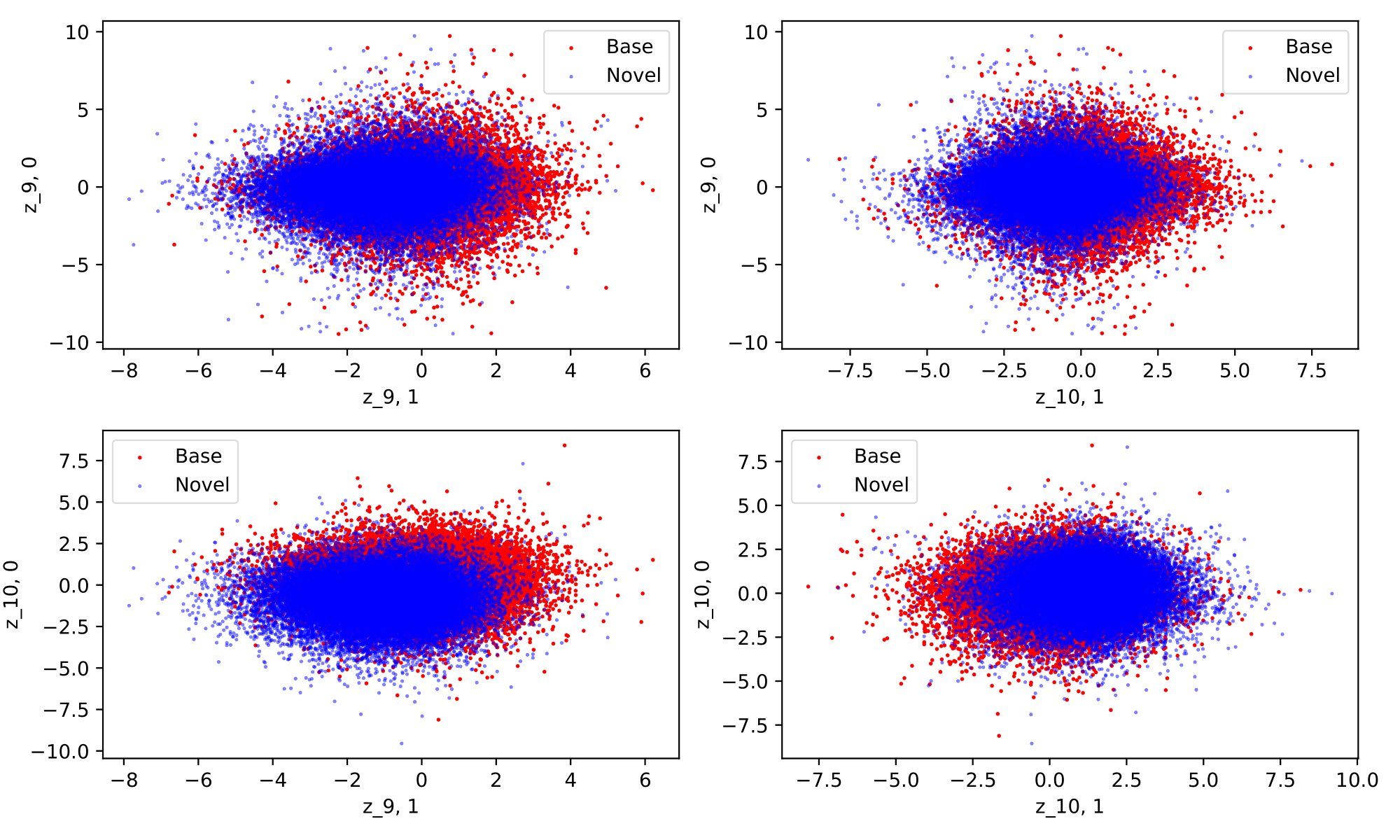}}
\caption{
Visualizations to substantiate our hypotheses on the Sth-Else dataset. Red: base data. Blue: novel data. 
(a) This UMap visualization contrasts the action features extracted using a static ViT-B/16 backbone (Radford et al., 2021) from the base and novel data within the Sth-Else dataset. The stark distributional disparities underscore the challenges inherent in few-shot action recognition learning. 
(b) A pairplot of UMap projections for the action feature embeddings, $\rvx$, and the latent variables, $\rvz$, derived from two instances of our model’s temporal dynamics generation module, trained separately on the base and novel data. 
(c) A pairplot comparing values of $\rvz_{n,t}$ from two models' temporal dynamics transition modules, each integrating the $\rvu$ from the base and novel data, respectively.  
(d) The temporal dynamics of latent variables using a consistent $\rvz^u_t$ across both base and novel data.
The close alignment observed across datasets implies that the temporal dynamics transition modules are consistent, confirming the stability of these functions regardless of the data source. 
}
\label{fig:motiv}
 \vspace{-0.3cm}
\end{figure}

  



\subsection{Discussion}\label{sec:discuss}

\bfsection{The distributional disparities between the base and novel data}
To illustrate the distributional differences between the base and novel data sets, we employed a pre-trained ViT-B/16 backbone to extract action representations from each.
As demonstrated in Figure~\ref{fig:motivating_a}, the Sth-Else dataset is characterized by significant distributional disparities between the base and novel data. This aspect notably intensifies the complexity of the few-shot action recognition task, highlighting the challenges inherent in adapting models from the base data to the novel data.

\bfsection{Validating of our assumptions}
To validate our theoretical assumptions, we independently trained two \ourmeos models on the base and novel data sets, respectively. 

We visualized several examples of the temporal dynamic transition module through pairplots representing the relationship between the latent variables across three consecutive time steps. We assume that if the temporal dynamic transition is consistent across the data, there should be a significant alignment between these variables. 
This alignment is indeed observed in Figure~\ref{fig:motivating_b} even under the severe data distributional disparities in Figure~\ref{fig:motivating_a}. Additionally, the impact of $\rvu$ on the temporal dynamics is shown in Figure~\ref{fig:motivating_d}. 
The better accuracy scores from Table~\ref{tab:sth-else} to ~\ref{tab:ablation} support our assumption that $\rvu$ captures the distributional discrepancies. 

Figure~\ref{fig:motivating_c} similarly displays the pairplots between $\rvx$ and $\rvz$ for all time steps, illustrating the relationship between observed data and latent variables, i.e., temporal dynamic generation. The apparent overlap suggests that our assumption that holding the temporal dynamic generation fixed holds true. 

We also empirically validate our assumptions on SSv2, HMDB-51, and UCF-101 datasets in Figures~\ref{fig:motiv_ssv2} in the appendix. We would like to highlight the consistency of the evidence regardless of the size of distributional disparities between the base and novel data.

\section{Conclusion}
\label{sec:conclusion}

We introduced the Domain-Invariant Temporal Dynamics (\ourmeo) framework for few-shot action recognition. Following the common two-stage training strategy, our framework involves unsupervised feature learning followed by supervised label prediction. In the unsupervised feature learning stage, we incorporate a learnable domain index to help the learned temporal dynamics modules to be more invariant to domain shift. These modules remain fixed during few-shot adaptation. Empirically, we validate the effectiveness of this assumption by achieving new state-of-the-art results on various standard action recognition datasets with minimal parameter updates during the adaptation process. 

\bfsection{Limitations}
This paper demonstrates the effectiveness of the learned temporal dynamics empirically. However, we believe that the effectiveness of any adaptation strategy from the base to novel data should be theoretically grounded, particularly concerning the generalization bound. Currently, this aspect is not adequately addressed in our work. Such an investigation could provide deeper insights into which components of a model should remain fixed versus which should be updated during adaptation.
Extending \ourmeos to address instantaneous causal relations and providing a generalization bound are clear directions for future work. 
Lastly, we mention that exploring the merits of \ourmeos for LLM agents might be an interesting direction~\citep{llmagent_arxiv13_1,llmagent_arxiv13_2}.

\section*{Acknowledgements}

This work is supported by National Science Foundation, Division of Information and Intelligent Systems (No. 2239688).

\section*{Impact Statement}

The challenge of recognizing human actions from video sequences is currently at the frontier of machine learning. It has many real-world applications ranging from security and behavior analysis to gaming and human-robot interaction. One of the major difficulties identified in the literature on action recognition is the perennial dearth of labeled data for certain action classes due to the sheer number of possible actions. Few-shot learning is one of the primary approaches to this problem of scarcity. However, few-shot learning struggles when there is a significant distribution shift between the original data used for the pre-trained model and the novel data from which one wishes to recognize actions. Our approach develops a model of temporal relations that distinguishes the parts of the model that transfer to the novel context from those that do not. By limiting the number of parameters that must be fine-tuned to the novel data we have developed a model that can be adapted efficiently and provides higher accuracy than the previous leading benchmarks. Our hope is that future work will extend our approach with increasingly realistic temporal relation models that enable action recognition with higher accuracy. While we recognize that the field of action recognition broadly does pose risks, including related to automated surveillance, our work does not present any significant new risk to the field. Ultimately, this paper presents work whose goal is to advance the field of Machine Learning, and particularly action recognition. There are many potential societal consequences of our work, none of which we feel must be specifically highlighted here.

\nocite{langley00}

\bibliography{example_paper}
\bibliographystyle{icml2024}

\newpage
\appendix
\onecolumn
\section{Transition Prior Likelihood Derivation}\label{ap:derive}

Consider a paradigmatic instance of latent causal processes. In this case, we are concerned with two time-delayed latent variables, namely, $\rvz_t = [\rvz_{1,t}, \rvz_{2,t}]$. Crucially, there is no inclusion of $\rvu$. We set time lag is defined as 1 for simplicity. This implies that each latent variable, $\rvz_{n,t}$, is formulated as $\rvz_{n,t}=f_n(\rvz_{t-1},\epsilon_{n,t})$, where the noise terms, $\epsilon_{n,t}$, are mutually independent. To represent this latent process more succinctly, we introduce a transformation map, denoted as $f$. It's worth noting that in this context, we employ an overloaded notation; specifically, the symbol $f$ serves dual purposes, representing both transition functions and the transformation map.

\begin{equation}
    \begin{bmatrix}
    \rvz_{1,t-1} \\
    \rvz_{2,t-1} \\
    \rvz_{1,t} \\
    \rvz_{2,t} \\
    \end{bmatrix} 
    =\mathbf{f} \left(
    \begin{bmatrix}
    \rvz_{1,t-1} \\
    \rvz_{2,t-1} \\
    \epsilon_{1,t} \\
    \epsilon_{2,t}
    \end{bmatrix}
    \right).
\end{equation}

By leveraging the change of variables formula on the map $\mathbf{f}$, we can evaluate the joint distribution of the latent variables $p(\rvz_{1,t-1}, \rvz_{2,t-1}, \rvz_{1,t},\rvz_{2,t})$ as:
\begin{equation}\label{eq:cvt-1}
p(\rvz_{1,t-1}, \rvz_{2,t-1}, \rvz_{1,t},\rvz_{2,t}) = p(\rvz_{1,t-1}, \rvz_{2,t-1}, \epsilon_{1,t},\epsilon_{2,t}) / \left|\det \mathbf{J}_\mathbf{f}\right|,
\end{equation}
where $\mathbf{J}_\mathbf{f}$ is the Jacobian matrix of the map $\mathbf{f}$, which is naturally a low-triangular matrix:
$$
\mathbf{J}_\mathbf{f} = 
\begin{bmatrix}
1 & 0 & 0 & 0\\
0 & 1 & 0 & 0\\
\frac{\partial \rvz_{1,t}}{\partial \rvz_{1,t-1}} & \frac{\partial \rvz_{1,t}}{\partial \rvz_{2,t-1}} & \frac{\partial \rvz_{1,t}}{\partial \epsilon_{1,t}} & 0 \\ 
\frac{\partial \rvz_{2,t}}{\partial \rvz_{1,t-1}} & \frac{\partial \rvz_{2,t}}{\partial \rvz_{2,t-1}} & 0 & \frac{\partial \rvz_{2,t}}{\partial \epsilon_{2,t}}
\end{bmatrix}.
$$
Given that this Jacobian is triangular, we can efficiently compute its determinant as $\prod_n \frac{\partial \rvz_{n,t}}{\partial \epsilon_{n,t}}$. Furthermore, because the noise terms are mutually independent, and hence $\epsilon_{n,t} \perp \epsilon_{l,t}$ for $m \neq n$ and $\epsilon_t \perp \rvz_{t-1}$, we can write Eq.~\ref{eq:cvt-1} as:

\begin{equation}\label{eq:example}
\begin{aligned}
    p(\rvz_{1,t-1}, \rvz_{2,t-1}, \rvz_{1,t},\rvz_{2,t}) &= p(\rvz_{1,t-1}, \rvz_{2,t-1}) \times p(\epsilon_{1,t},\epsilon_{2,t}) / \left|\det \mathbf{J}_\mathbf{f}\right| \quad (\text{because }\epsilon_t \perp \rvz_{t-1})\\
    &= p(\rvz_{1,t-1}, \rvz_{2,t-1}) \times \prod_n p(\epsilon_{n,t}) / \left|\det \mathbf{J}_\mathbf{f}\right| \quad (\text{because }\epsilon_{1,t} \perp \epsilon_{2,t})
\end{aligned}    
\end{equation}

By eliminating the marginals of the lagged latent variable $p(\rvz_{1,t-1}, \rvz_{2,t-1})$ on both sides, we derive the transition prior likelihood as:

\begin{equation}\label{eq:example-likelihood}
p( \rvz_{1,t},\rvz_{2,t} \vert \rvz_{1,t-1}, \rvz_{2,t-1}) = \prod_n p(\epsilon_{n,t}) / \left|\det \mathbf{J}_\mathbf{f}\right| = \prod_n p(\epsilon_{n,t}) \times \left|\det \mathbf{J}_\mathbf{f}^{-1}\right|.
\end{equation}


Let $\{f^{-1}_{n}\}_{n=1,2,3...}$ be a set of learned inverse dynamics transition functions that take the estimated latent causal variables in the fixed dynamics subspace and lagged latent variables, and output the noise terms, i.e., $\hat{\epsilon}_{n,t} = f^{-1}_{n}\left(\hat{\rvz}_{n,t}, \text{Pa}(\hat{\rvz}_{n,t}) \right)$. 

The differences of our model from Eq.~\ref{eq:example-likelihood} are that the learned inverse dynamics transition functions take $\rvz^u_t$ as input arguments to out the noise terms, i.e, $\hat{\epsilon}_{n,t} = f^{-1}_{n}\left(\hat{\rvz}_{n,t}, \text{Pa}(\hat{\rvz}_{n,t}),  \rvu_t \right)$. 
\begin{align}
\log p\left(\hat{\rvz}_t \vert \hat{\rvz}_{t-l}, \rvz^u_t\right) = \sum_{n=1}^N \log p(\hat{\epsilon}_{n,t} \vert \rvz^u_t)+ \sum_{n=1}^N \log \Big| \frac{\partial f^{-1}_{n}}{\partial \hat{\rvz}_{n,t}}\Big|
\end{align}

\section{Assumption validation on SSv2, HMDB51 and UCF101 dataset}

\begin{figure}[t]
\captionsetup[subfloat]{labelformat=simple}
\subfloat[\label{fig:motivating_ssv2_a}]{
\includegraphics[width=0.23\linewidth]{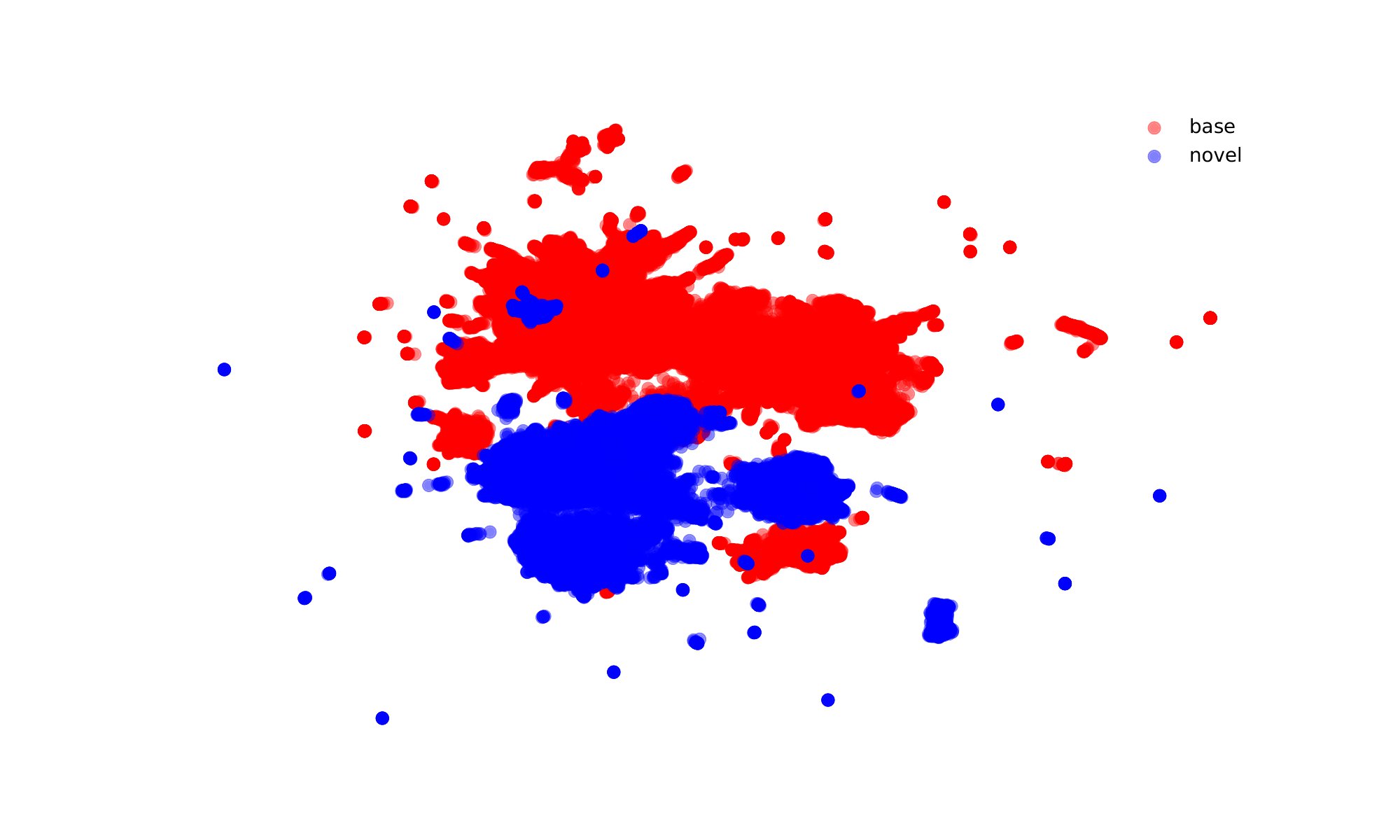}}
\subfloat[\label{fig:motivating_ssv2_c}]{
\includegraphics[width=0.23\linewidth]
{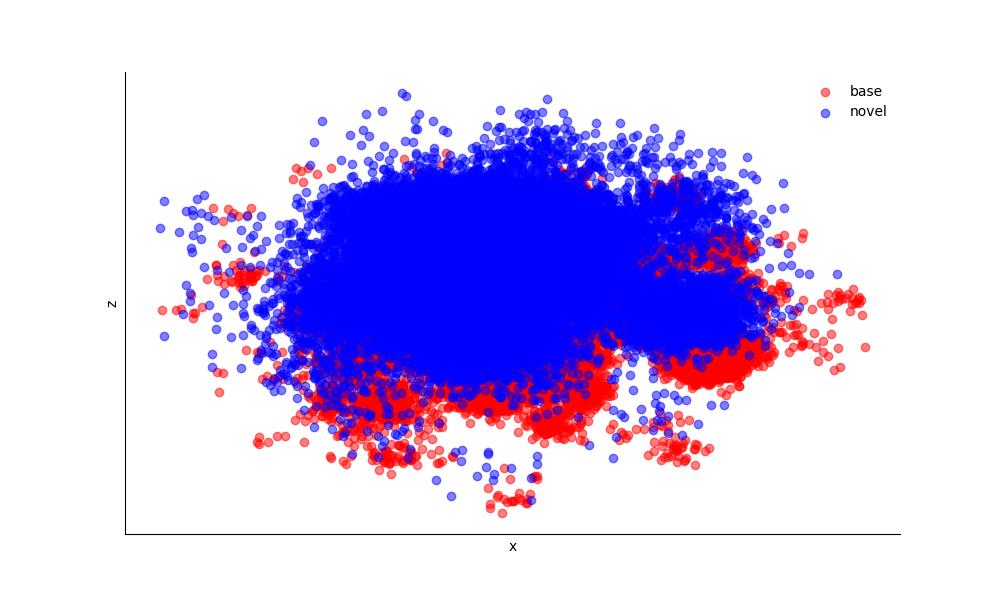}}
\subfloat[\label{fig:motivating_ssv2_b}]{
\includegraphics[width=0.26\linewidth]
{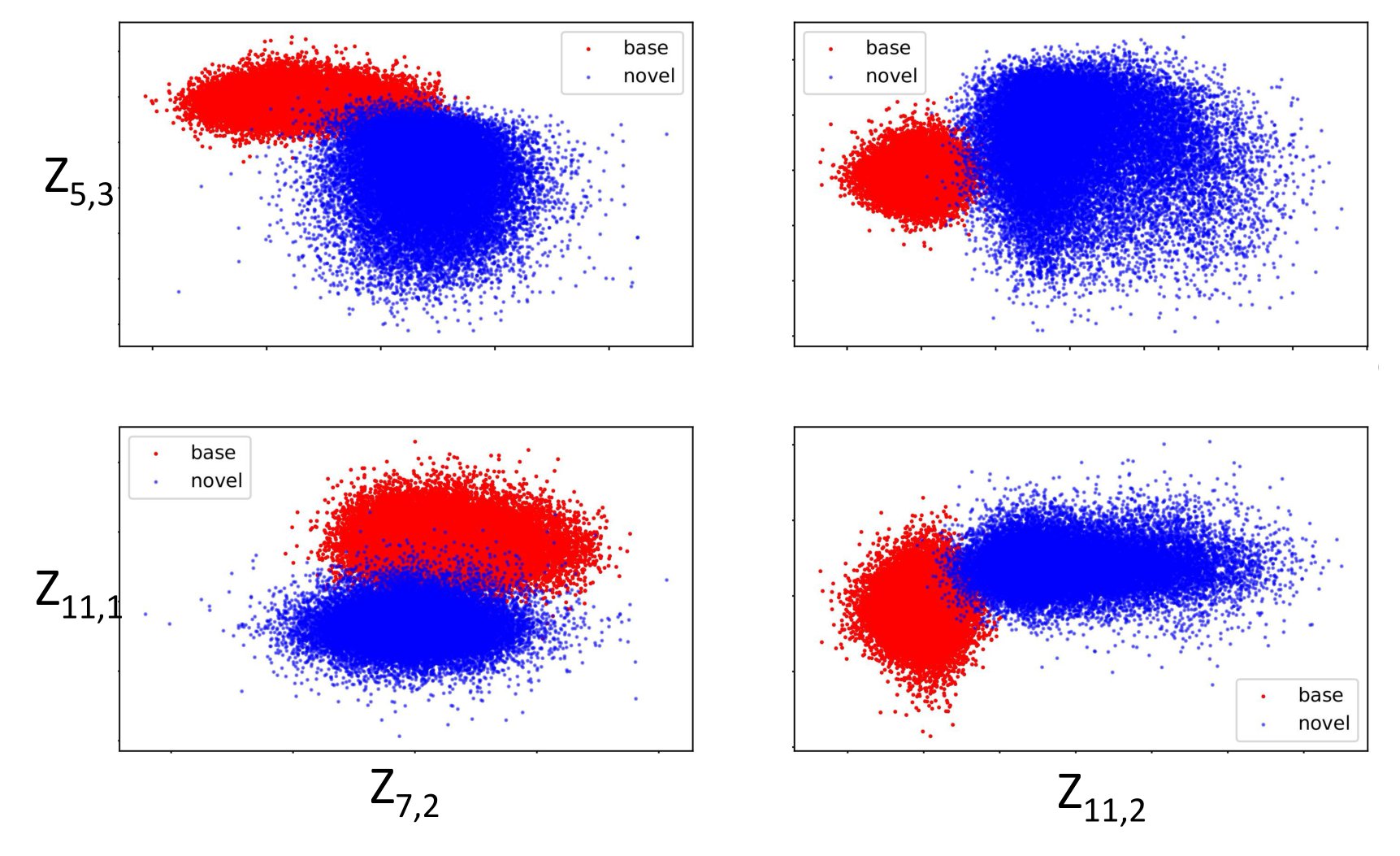}}
\subfloat[\label{fig:motivating_ssv2_d}]{
\includegraphics[width=0.26\linewidth]
{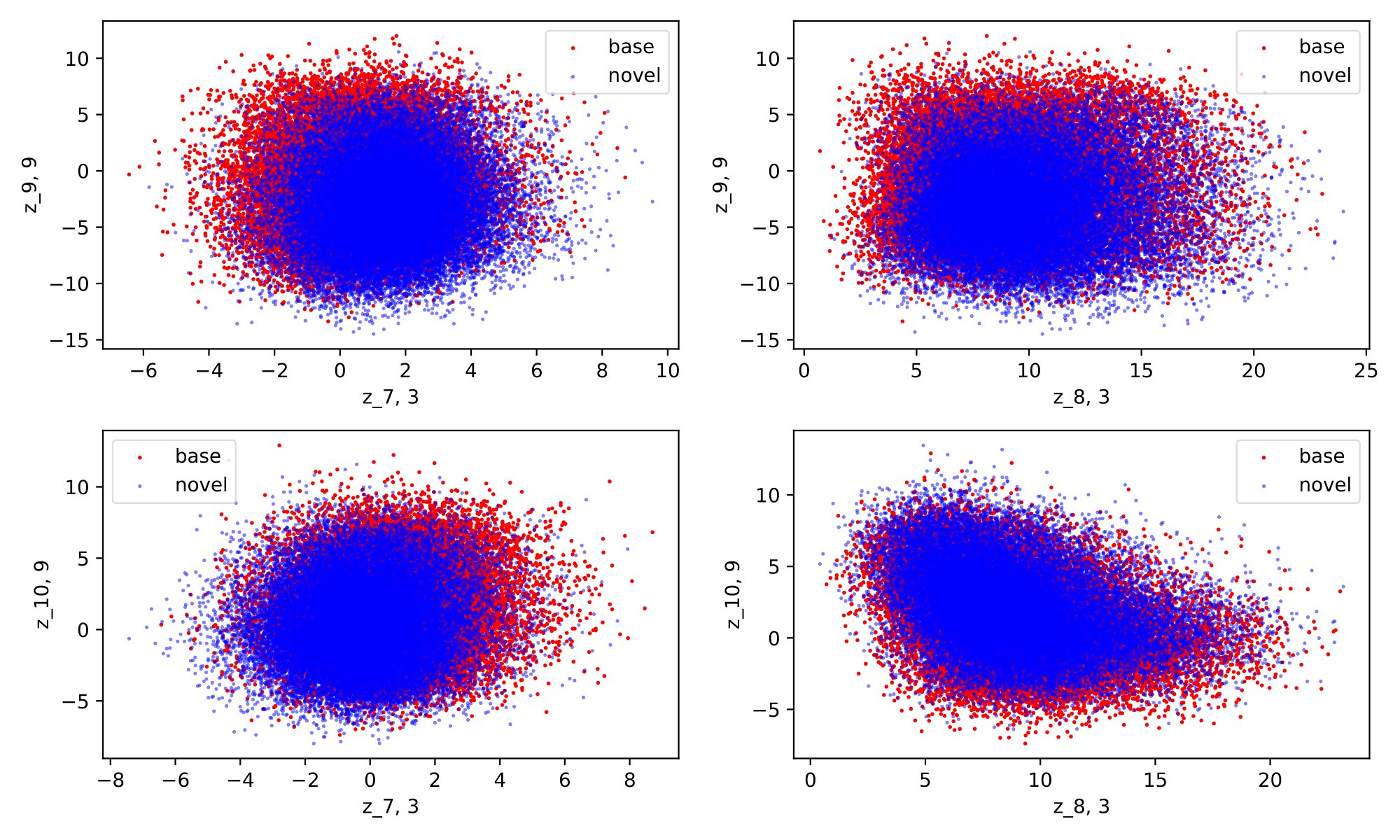}}\\
\subfloat[\label{fig:motivating_hmdb_a}]{
\includegraphics[width=0.23\linewidth]{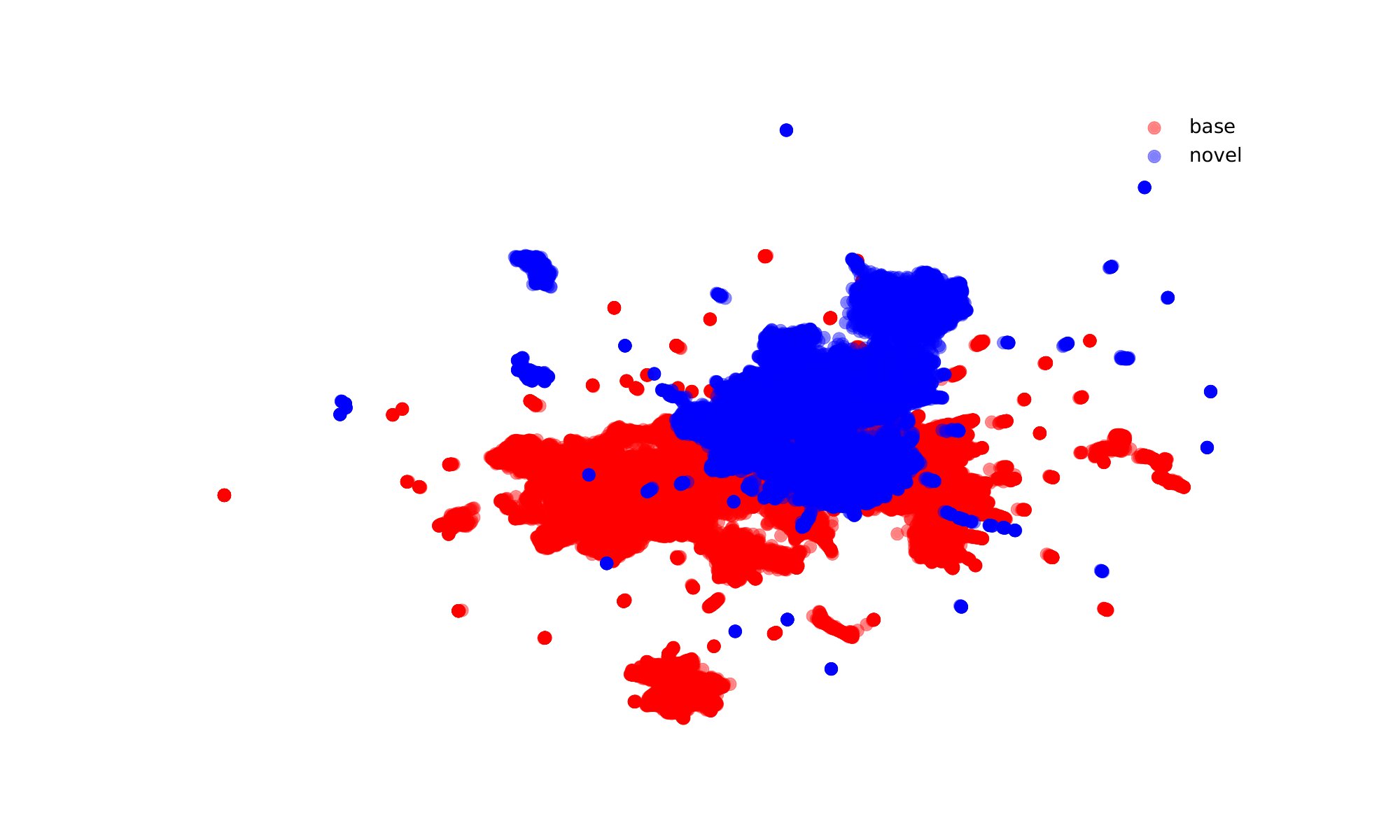}}
\subfloat[\label{fig:motivating_hmdb_c}]{
\includegraphics[width=0.23\linewidth]
{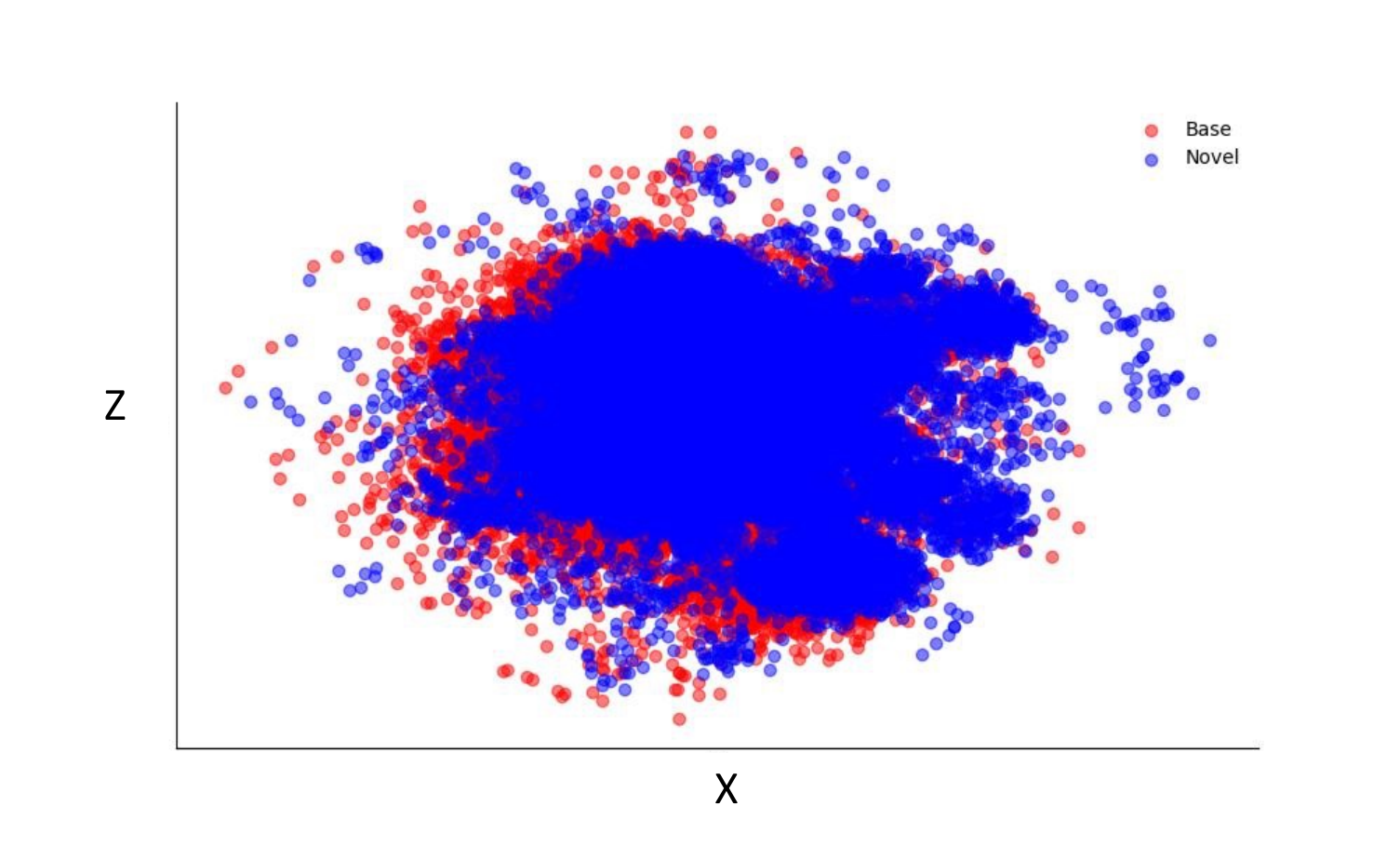}}
\subfloat[\label{fig:motivating_hmdb_b}]{
\includegraphics[width=0.25\linewidth]
{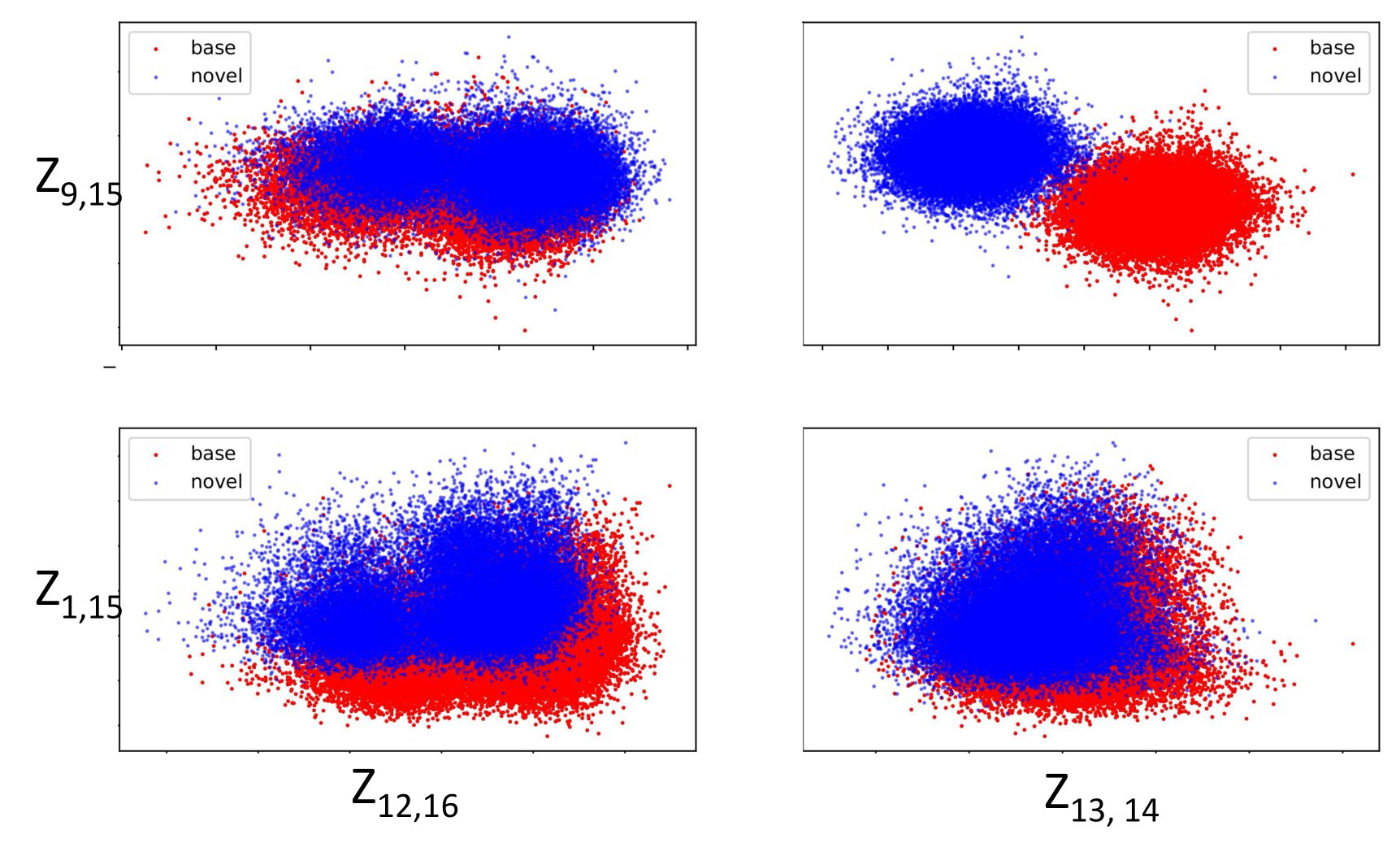}}
\subfloat[\label{fig:motivating_hmdb_d}]{
\includegraphics[width=0.25\linewidth]
{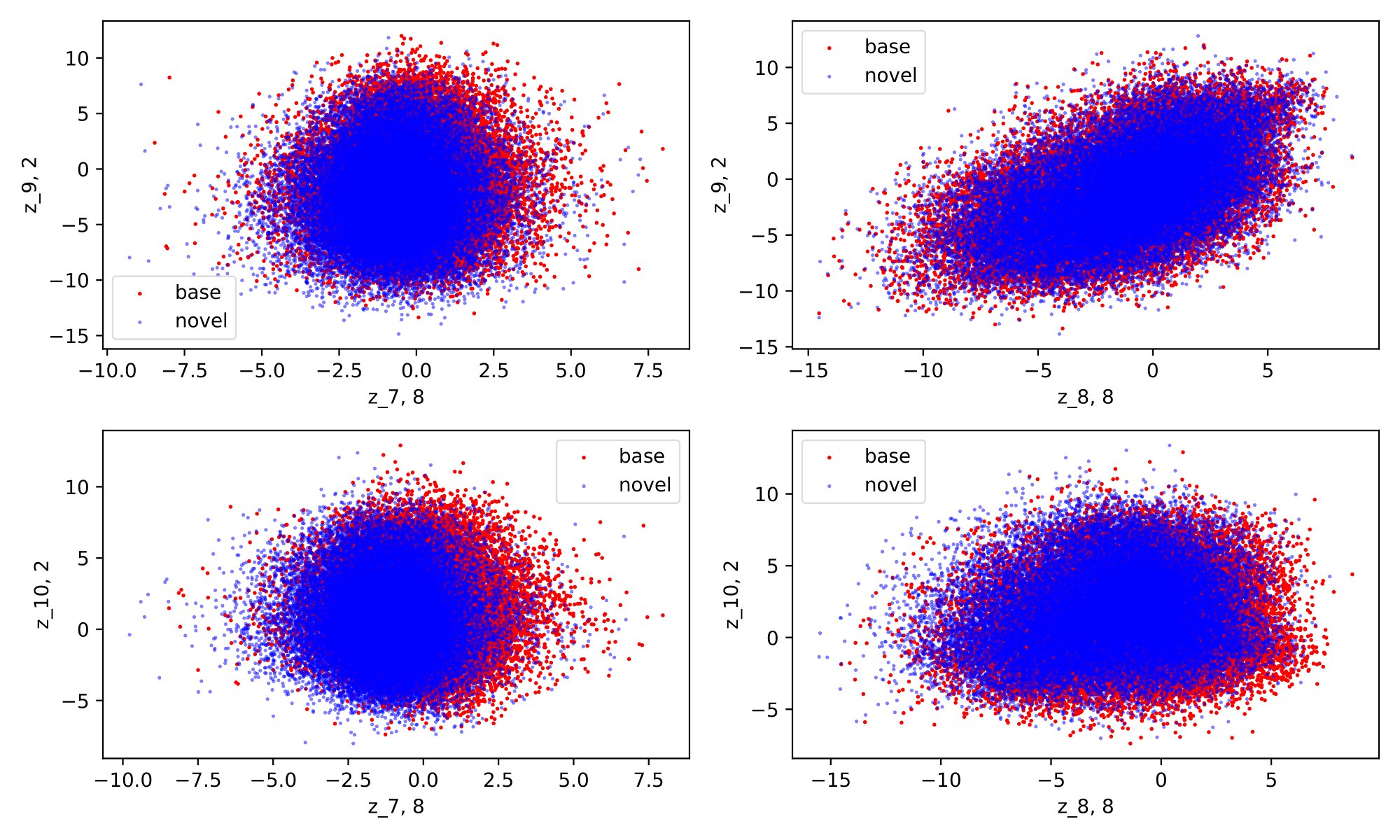}}\\
\subfloat[\label{fig:motivating_ucf_a}]{
\includegraphics[width=0.23\linewidth]
{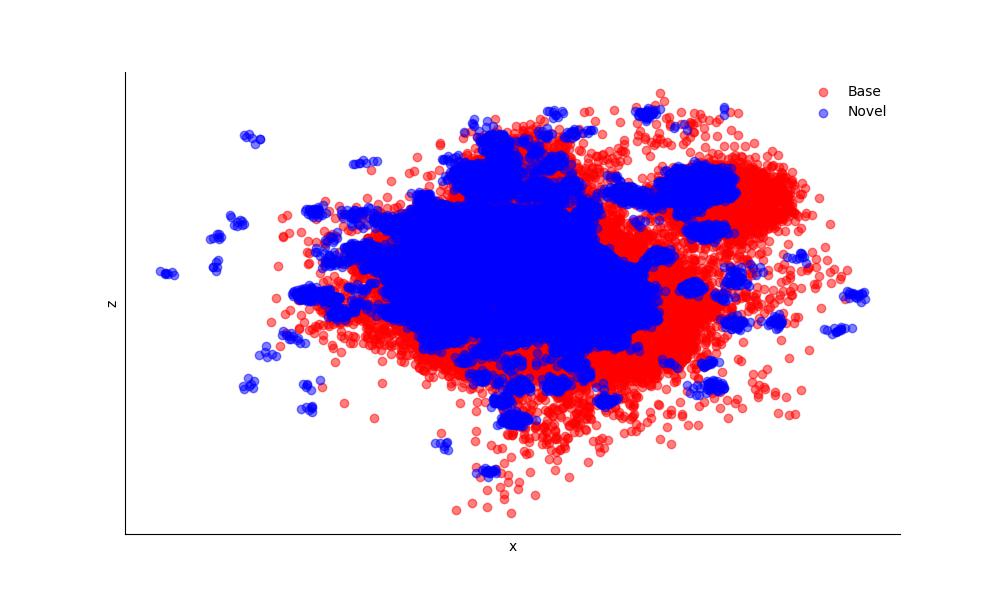}}
\subfloat[\label{fig:motivating_ucf_c}]{
\includegraphics[width=0.23\linewidth]
{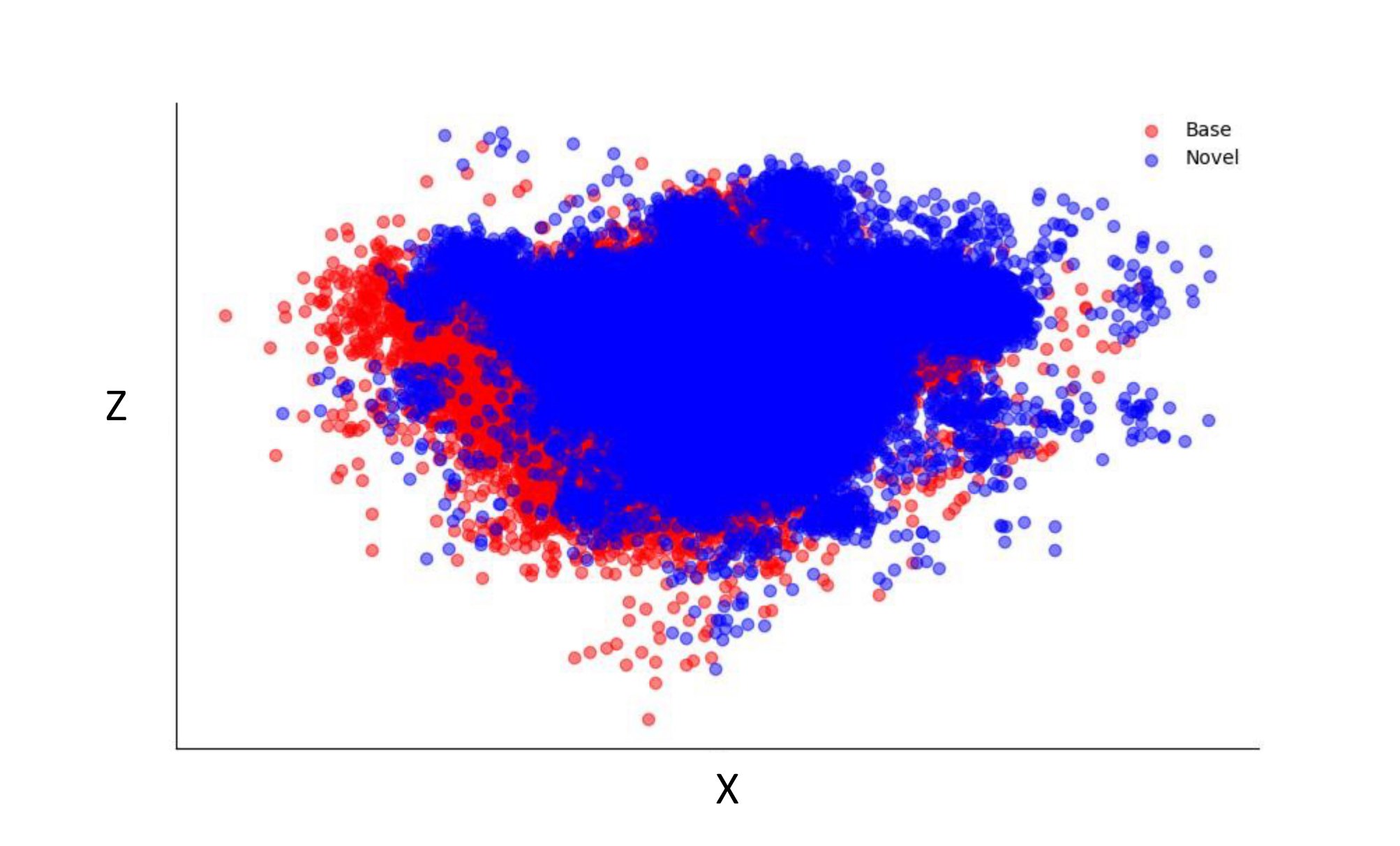}}
\subfloat[\label{fig:motivating_ucf_b}]{
\includegraphics[width=0.25\linewidth]
{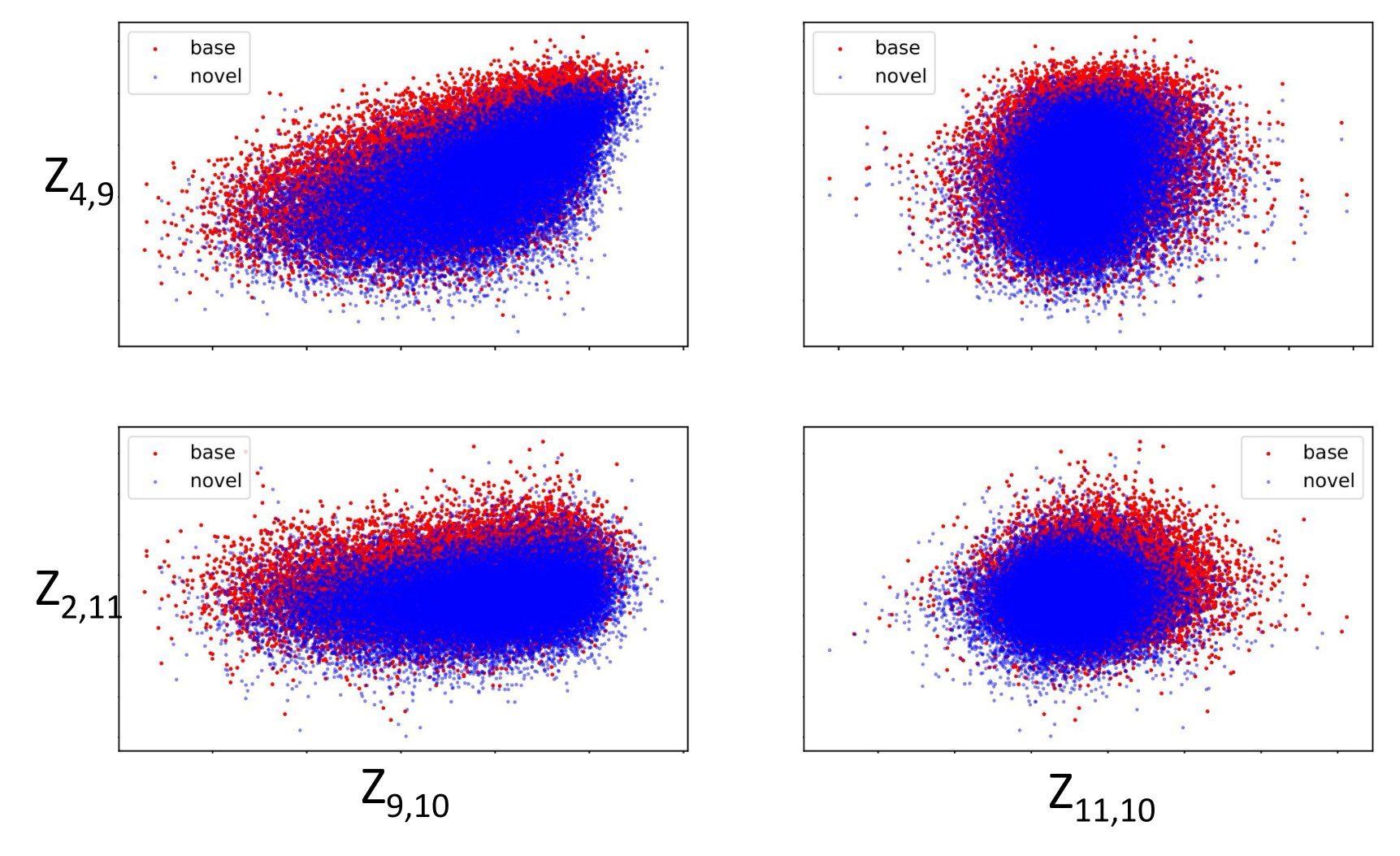}}
\subfloat[\label{fig:motivating_ucf_d}]{
\includegraphics[width=0.25\linewidth]
{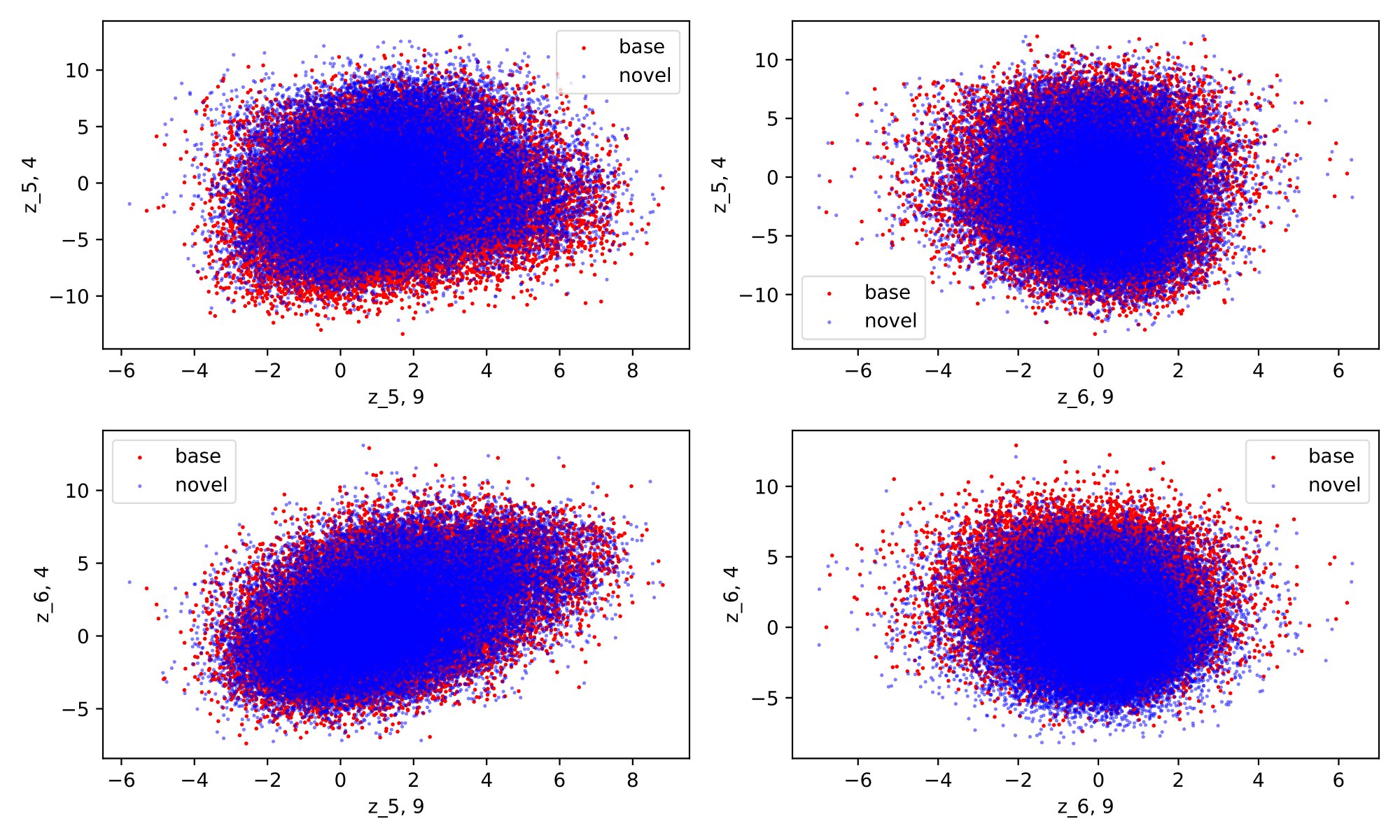}}
\caption{ \small
Visualizations validating hypotheses on SSv2, HMDB-51 and UCF-101 datasets (from top row to bottom row), respectively. Red: base data (K-400), blue: novel data.
(a,e,i) UMap of action embeddings.
(b,f,j) Pairplots of UMap projections for embeddings and latent variables.
(c,g,k) Pairplots of latent variables from temporal invariance functions.
(d,h,l) Temporal dynamics of latent variables with consistent $\rvz^u_t$.
}
\label{fig:motiv_ssv2}
\vspace{-0.2cm}
\end{figure}

Figure~\ref{fig:motiv_ssv2} showcases the similar motivating examples of a few-shot action recognition task with \melater in using SSv2, HMDB-51 and UCF-101 as novel data, while the K-400 serves as base data. Please see the caption for the detailed explanations.

\section{Visual results on the Sth-Else dataset}

\begin{figure}[h!]
\begin{center}
\includegraphics[width=0.49\linewidth]
{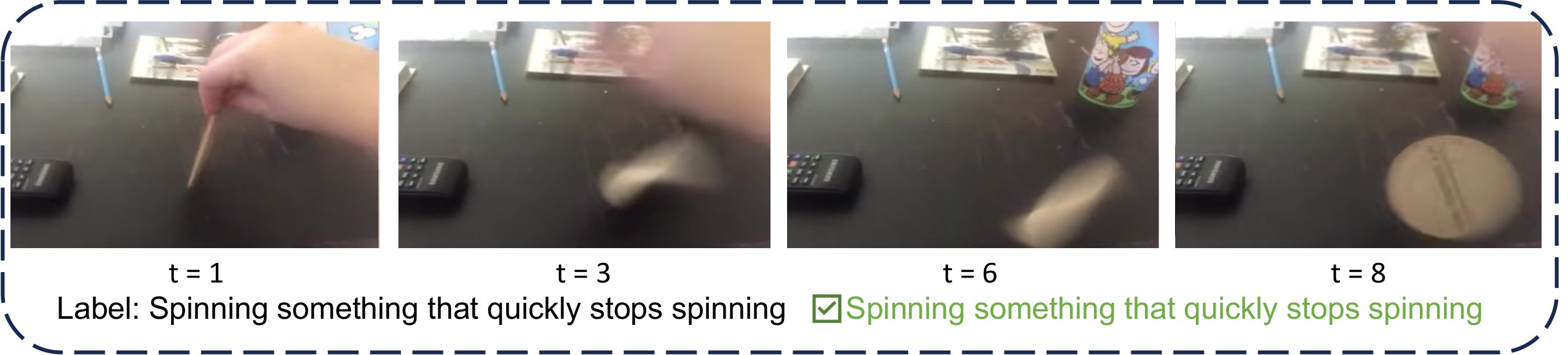}
\includegraphics[width=0.49\linewidth]
{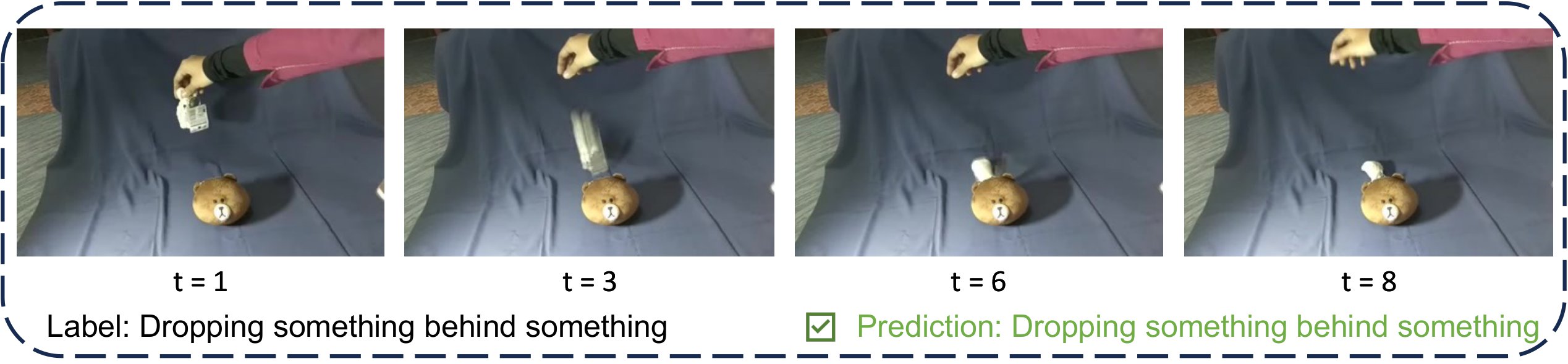}\\
\includegraphics[width=0.49\linewidth]
{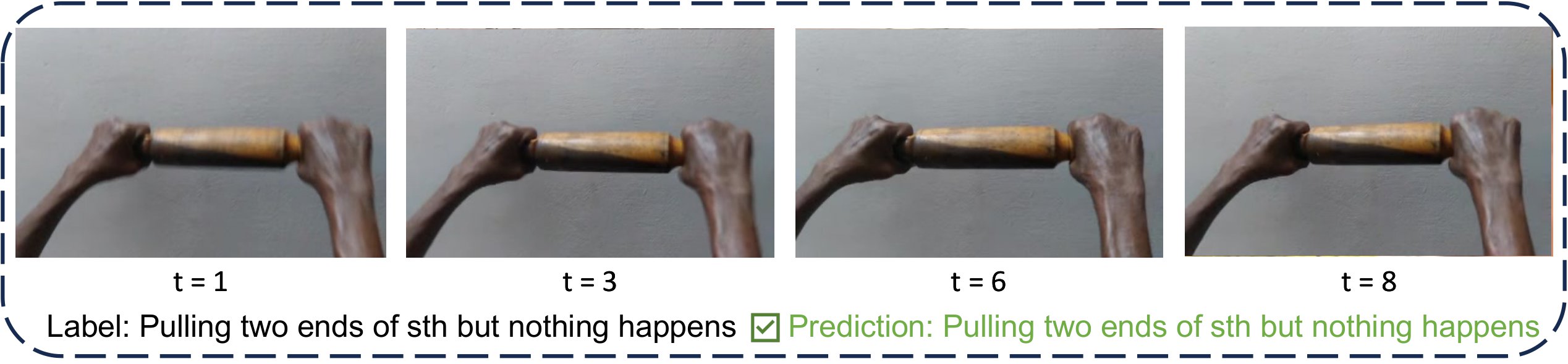}
\includegraphics[width=0.49\linewidth]
{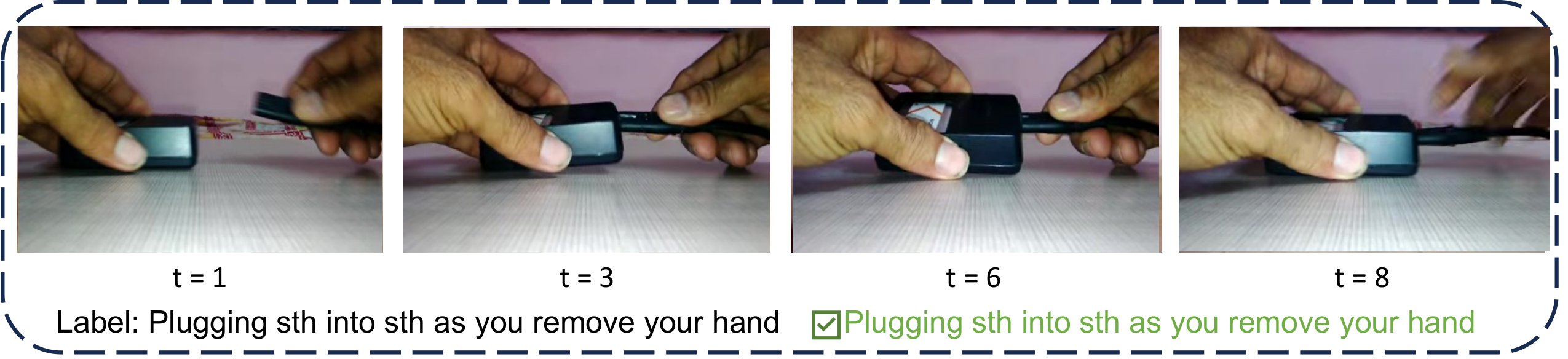}\\
\includegraphics[width=0.49\linewidth]
{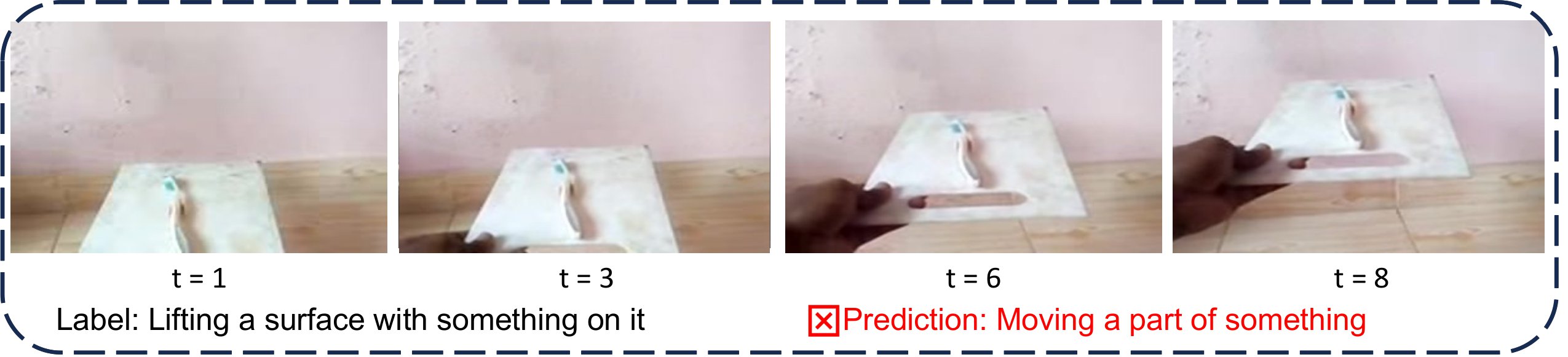}
\includegraphics[width=0.49\linewidth]
{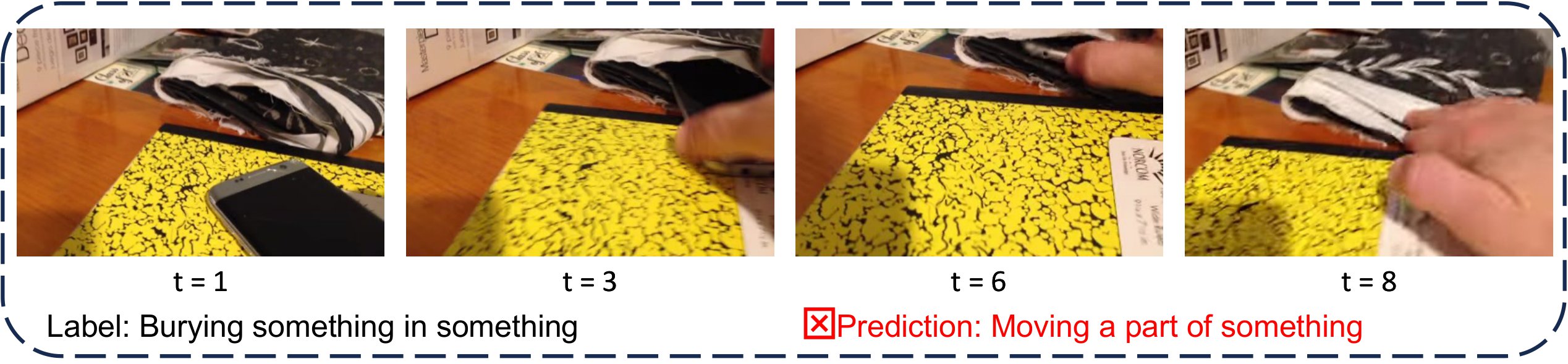}\\
\end{center}
\caption{The visual examples of \ourmeo$_{NCE}$ on Sth-Else dataset. \ourmeo$_{NCE}$ correctly predicts the four actions but misclassifies two videos of ``lifting a surface with something on it'', and ``Burying something in something'' as ``Moving a part of something''}
\label{fig:qr}
\end{figure}

Fig.~\ref{fig:qr} shows a few examples of action recognition results of our \ourmeo\HS on the Sth-Else dataset.

\section{Additional Experiments}
In this section, we focus on comparisons with state-of-the-art metric-based methods, including MoLo \citep{few_cvpr23}, HySRM \citep{few_cvpr22_2}, HCL \citep{hcl_eccv22},  OTAM \citep{few_cvpr20}, TRX \citep{few_cvpr21} and STRM \citep{few_cvpr22_1}. For a fair evaluation, we perform experiments across the SSv2, UCF-101, HMDB-51, and Kinetics datasets.
In the SSv2-Full and SSv2-Small datasets, we randomly selected 64 classes for $\mathcal{D}$ and 24 for $\mathcal{S}$ and $\mathcal{Q}$. The main difference between SSv2-Full and SSv2-Small is the dataset size, with SSv2-Full containing all samples per category and SSv2-Small including only 100 samples per category. For HMDB-51, we chose 31 action categories for $\mathcal{D}$ and 10 for $\mathcal{S}$ and $\mathcal{Q}$, while for UCF-101, the selection was 70 and 21 categories, respectively. For Kinect, we used 64 action categories for $\mathcal{D}$ and 24 for $\mathcal{S}$ and $\mathcal{Q}$. To maintain statistical significance, we executed 200 trials, each involving random samplings across categories. After training on $\mathcal{D}$, we used $k$ video sequences from each action category to form $\mathcal{S}$ for model updates. The inference phase utilized the remaining data from $\mathcal{Q}$.

Table~\ref{tab:nway_p1} and Table~\ref{tab:nway_p2} show additional experiments for 5-way-k-shot learning where the base and novel data are taken from the same original dataset. Between the two tables we observe that \ourmeo$_{CE}$ achieves the highest Top-1 accuracy in 11 out of 12 experiments, coming in second by only $0.4$ with $k=5$ on the UCF-101 dataset.
\begin{table}[h!]
\small
\caption{\small Comparing \ourmeo$_{CE}$ to benchmarks for 5-way-k-shot learning on the SSv2 and SSv2-small using the ResNet-50 backbone}
\vspace{-0.4cm}
\label{tab:nway_p1}
\begin{center}
\begin{tabular}{l|ccc|ccc}
\hline\hline
{}
&\multicolumn{3}{c}{SSv2}
&\multicolumn{3}{c}{SSv2-small}
\\ \hline
{}
&{1-shot}
&{3-shot}
&{5-shot}
&{1-shot}
&{3-shot}
&{5-shot}
\\ \hline 
OTAM  & 42.8 & 51.5 & 52.3 & 36.4 & 45.9 & 48.0\\
TRX  & 42.0 & 57.6 & 62.6 & 36.0 & 51.9 & 56.7\\
STRM  & 42.0 & 59.1 & 68.1 & 37.1 & 49.2 & 55.3\\
HyRSM  & 54.3 & 65.1 & 69.0 & 40.6 & 52.3 & 56.1\\
HCL  & 47.3 & 59.0 & 64.9 & 38.7 & 49.1 & 55.4\\
MoLo  & \underline{56.6} & \underline{67.0} & \underline{70.6} & \underline{42.7} & \underline{52.9} & \underline{56.4}\\
\bf{\ourmeo$_{CE}$}  & \bf{60.0} & \bf{68.3} & \bf{71.9} & \bf{45.8} & \bf{53.6} & \bf{58.0}\\
\hline\hline
\end{tabular}
\end{center}
\end{table}
\begin{table}[h!]
\small
\caption{\small Comparing \ourmeo$_{CE}$ to benchmarks for 5-way-k-shot learning on the UCF-101, HMDB-51, and Kinectics datasets using the ResNet-50 backbone.}
\vspace{-0.4cm}
\label{tab:nway_p2}
\begin{center}
\begin{tabular}{l|cc|cc|cc}
\hline\hline
{}
&\multicolumn{2}{c}{UCF-101}
&\multicolumn{2}{c}{HMDB-51}
&\multicolumn{2}{c}{Kinects}
\\ \hline
{}
&{1-shot}
&{5-shot}
&{1-shot}
&{5-shot}
&{1-shot}
&{5-shot}
\\ \hline 
OTAM & 79.9 & 88.9 & 54.5 & 68.0 & 79.9 & 88.9\\
TRX  & 78.2 & 96.1 & 53.1 & 75.6 & 78.2 & 96.2\\
STRM  & 80.5 & \bf{96.9} & 52.3 & 77.3 & 80.5 & \underline{96.9}\\
HyRSM  & 83.9 & 94.7 & 60.3 & 76.0 & 83.9 & 94.7\\
HCL  & 82.8 & 93.3 & 59.1 & 76.3 & 73.7 & 85.8\\
MoLo  & \underline{86.0} & 95.5 & \underline{60.8} & \underline{77.4} & \underline{86.0} & 95.5\\
\bf{\ourmeo$_{CE}$}  & \bf{87.3} & \underline{96.5} & \bf{61.9} & \bf{80.5} & \bf{86.1} & \bf{98.0}\\
\hline\hline
\end{tabular}
\end{center}
\end{table}

\subsection{Comparisons to Augmented Models}
We further assess if adding $\rvu$ to other methods would further improve the results in Table~\ref{tab:sthelse_supp}. The results indicate marginal improvements over the original methods. Again, our \melater obtains the highest accuracy among these methods.

\begin{table}[h]
\small
\caption{\small Additional comparisons by augmenting existing methods on the Sth-Else dataset using the ViT-B/16 backbone. $+\rvu$ means the method updates the domain encoder when adapting instead of fine-tuning. Since VL Prompting uses VPT \citep{vpt_eccv22} within the ViFi-CLIP framework, we only test ViFi-CLIP$+\rvu$.}
\vspace{-0.4cm}
\label{tab:sthelse_supp}
\begin{center}
\begin{tabular}{l|cc}
\hline\hline
{}
&\multicolumn{2}{c}{Sth-Else}
\\ \hline
{}
&{5-shot}
&{10-shot}
\\ \hline 
ORViT  & 33.3 & 40.2 \\
ORViT + $\rvu$  & 33.9 & 41.8 \\
SViT  & 34.4 & \underline{42.6}\\
SViT + $\rvu$  & \underline{35.2} & \bf{44.0}\\
\bf{\ourmeo$_{CE}$}  & \bf{37.6} & \bf{44.0} \\
\hdashline
ViFi-CLIP  & 44.5 & 54.0 \\
VL Prompting  & 44.9 & \underline{58.2} \\
ViFi-CLIP + $\rvu$  & \underline{45.2} & 58.0 \\
\bf{\ourmeo$_{NCE}$}  & \bf{48.5} & \bf{59.9} \\
\hline\hline
\end{tabular}
\vspace{-0.4cm}
\end{center}
\end{table}

\section{Network Architectures} 

Tab.~\ref{tab:architecture} illustrates the details of our implementation on \ourmeo$_{NCE}$.

\begin{table}[h]
\caption{The details of our network architectures for \ourmeo$_{NCE}$, where BS means batch size.}
\label{tab:architecture}
\begin{center}
\begin{tabular}{lll}
\multicolumn{1}{c}{Configuration}  
&\multicolumn{1}{c}{Description}
&\multicolumn{1}{c}{Output dimensions}
\\\hline \\
Image encoder       &   & \\
Input: $\text{concat}(\rvx_{1:T})$  &  & $\text{BS}\times T\times 1024$ \\
Dense &  256 neurons, LeakyReLU & $\text{BS}\times T\times 256$\\
Dense &  256 neurons, LeakyReLU & $\text{BS}\times T\times 256$\\
Dense &  Temporal embeddings & $\text{BS}\times T\times 2N$\\
Bottleneck &  Compute mean and variance of posterior & $\mu$, $\sigma$\\
Reparameterization &  Sequential sampling & $\hat{\rvz}_{1:T}$\\
\hline \\
Temporal dynamic
generation        &  \\
Input: $\hat{\rvz}_{1:T}$ &   & $\text{BS}\times T\times N$\\
Dense &  256 neurons, LeakyReLU & $\text{BS}\times T\times 256$\\
Dense &  256 neurons, LeakyReLU & $\text{BS}\times T\times 256$\\
Dense &  input embeddings & $\text{BS}\times T\times 1024$\\
\hline \\
Temporal dynamic
transition       &  \\
Input &  $\hat{\rvz}_{1:T}$ & $\text{BS}\times T\times N$\\
InverseTransition &  $\epsilon_{t}$ & $\text{BS}\times T\times N$\\
JacobianCompute &  $\log\text{det}|J|$ & $\text{BS}$\\
\hline \\
Classifier       &  \\
Input: $\text{Concat}(\hat{\rvz}_{1:T})$ &  & $\text{BS}\times T\times N$\\
Dense &  256 neurons, LeakyReLU & $\text{BS}\times T\times 256$\\
Dense &  256 neurons, LeakyReLU & $\text{BS}\times T\times 256$\\
Dense &  output embeddings & $\text{BS}\times T\times 1024$\\
\hline
\hline
\end{tabular}
\end{center}
\end{table}

\end{document}